\documentclass{article} 
\usepackage{iclr2027_conference,times}

\usepackage{amsmath,amsfonts,bm}

\def\eqref#1{equation~\ref{#1}}

\def\1{\bm{1}}

\DeclareMathAlphabet{\mathsfit}{\encodingdefault}{\sfdefault}{m}{sl}
\SetMathAlphabet{\mathsfit}{bold}{\encodingdefault}{\sfdefault}{bx}{n}

\usepackage{multirow}
\usepackage{hyperref}
\usepackage{url}
\usepackage{amssymb}
\usepackage{booktabs}
\usepackage{graphicx}
\usepackage[table]{xcolor}
\usepackage{enumitem}
\usepackage{graphicx}
\usepackage{subcaption}
\usepackage{wrapfig}

\definecolor{cellgreen}{HTML}{B3E0C8}
\definecolor{cellred}{HTML}{EEADAF}
\newcommand{\cellup}[2]{%
  \ifdim #2pt<1pt
    \cellcolor{cellgreen!25!white}{#1{\scriptsize\,($\uparrow$#2)}}%
  \else\ifdim #2pt<3pt
    \cellcolor{cellgreen!50!white}{#1{\scriptsize\,($\uparrow$#2)}}%
  \else\ifdim #2pt<6pt
    \cellcolor{cellgreen!75!white}{#1{\scriptsize\,($\uparrow$#2)}}%
  \else
    \cellcolor{cellgreen!100!white}{#1{\scriptsize\,($\uparrow$#2)}}%
  \fi\fi\fi
}

\newcommand{\celldown}[2]{%
  \ifdim #2pt<1pt
    \cellcolor{cellred!25!white}{#1{\scriptsize\,($\downarrow$#2)}}%
  \else\ifdim #2pt<3pt
    \cellcolor{cellred!50!white}{#1{\scriptsize\,($\downarrow$#2)}}%
  \else\ifdim #2pt<6pt
    \cellcolor{cellred!75!white}{#1{\scriptsize\,($\downarrow$#2)}}%
  \else
    \cellcolor{cellred!100!white}{#1{\scriptsize\,($\downarrow$#2)}}%
  \fi\fi\fi
}

\newcommand{\cellzero}[1]{%
  \cellcolor{gray!8}{#1{\scriptsize\,($\leftrightarrow$0.0)}}%
}

\usepackage[most]{tcolorbox}

\definecolor{takeawaybg}{RGB}{224,223,255}
\newtcolorbox{safetakeaways}{
  width=\columnwidth,
  colback=takeawaybg,
  colframe=takeawaybg,
  boxsep=0pt,
  left=10pt,
  right=10pt,
  top=2pt,
  bottom=3pt,
  fontupper=\linespread{0.9}\selectfont
}

\title{When Safety Rationales Hallucinate: \\ Diagnosing and Grounding Multimodal \\ Safety Guards}

\iclrfinalcopy
\author{%
  Yingshuo Wang\textsuperscript{1}, Weixiang Zhao\textsuperscript{1}, Guancheng Li\textsuperscript{1}, Jiahe Guo\textsuperscript{1}, Yu Zhang\textsuperscript{1}, Yanyan Zhao\textsuperscript{1} \\ 
  \textsuperscript{1}Harbin Institute of Technology\\
  \texttt{\{yswang,wxzhao,gcli,jhguo,yzhang,yyzhao\}@ir.hit.edu.cn} \\
}

\begin{document}

\maketitle

\begin{abstract}
Multimodal safety guards increasingly generate rationales alongside safety predictions, yet whether these rationales are reliable remains largely unexplored. We systematically study \emph{rationale hallucination} from two perspectives: visual rationale hallucination, where rationales contain unsupported visual facts, and safety attribution hallucination, where grounded content is assigned an incorrect safety interpretation. Across rationale-generation methods, multiple vision-language backbones, and safety datasets, we find that rationale hallucination is widespread and does not consistently diminish with higher classification accuracy or increased model scale. Further analysis of the Think-with-Image paradigm shows that hallucination is concentrated in fine-grained attributes and relations, while visual tools are often invoked infrequently or without sufficient spatial focus. Based on these findings, we propose \emph{Tool-Contrastive Grounding Reinforcement Learning} (TCG-RL), which rewards tool use only when it improves rationale grounding over a tool-free reference. TCG-RL substantially reduces rationale hallucination while maintaining competitive safety prediction performance. By revealing the gap between prediction correctness and rationale reliability, our study motivates a shift in multimodal safety guarding from outcome-only evaluation toward safety decisions that are both accurate and verifiably grounded.
\end{abstract}

\begin{center}
  \textit{\textcolor{red}{This paper contains demonstration content that may be disturbing for some readers.}}
\end{center}

\section{Introduction}
\label{introduction}

As multimodal foundation models are increasingly deployed in real-world applications, multimodal safety guard models have become an important component for identifying potentially harmful image-text content~\citep{safebench,mmsafetybench,vlguard,llavaguard,rjudge,safetygeometry}. Early studies mainly formulate multimodal safety guarding as a direct detection problem, where models output a binary label indicating whether an input is harmful or benign~\citep{harmeme,mami,pridemm}. More recent work has gradually extended this paradigm by requiring models to generate rationales that explain the basis of their safety decisions~\citep{memeintel,mememind,expohm,thinkingmeme,hmguguard}. Such rationales can make multimodal safety decisions more interpretable, traceable, and verifiable, especially when potential risks arise from the image, the text, or their implicit relation.

\begin{figure*}[t]
    \centering
    \includegraphics[width=\textwidth]{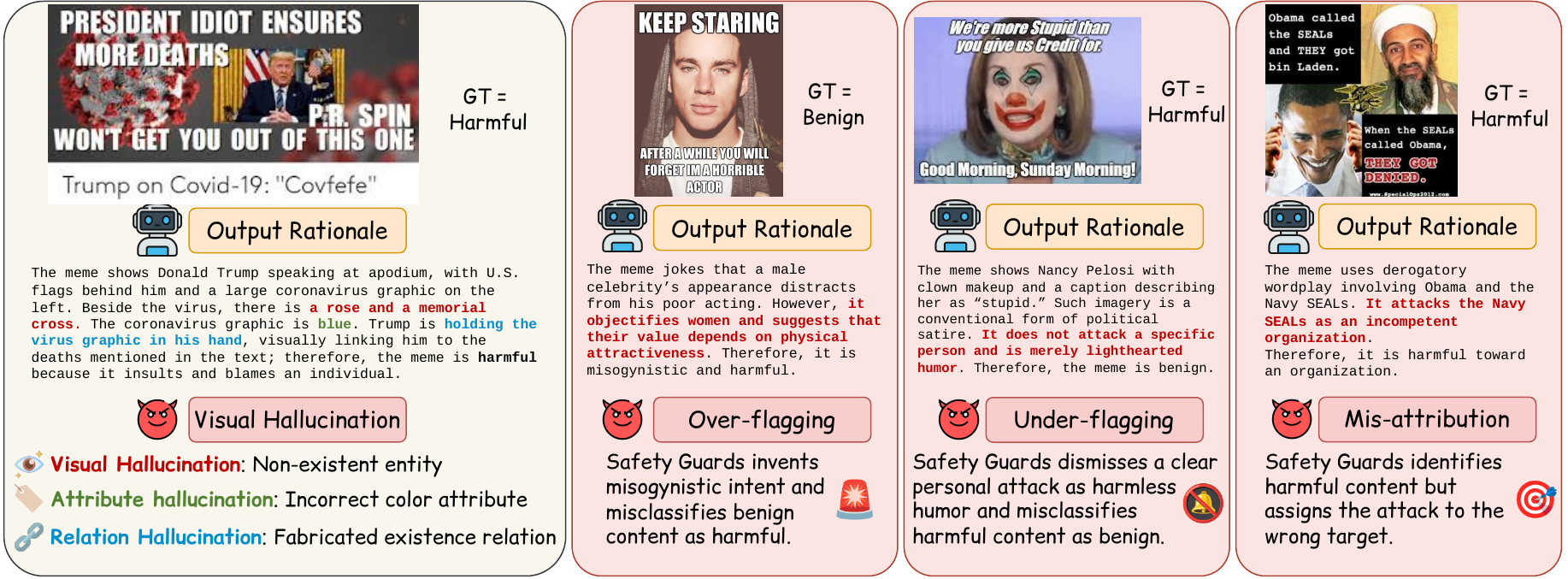}
    \caption{Examples of rationale hallucination in multimodal safety guards. The left panel
shows visual evidence hallucination caused by unsupported entities, attributes,
and relations. The three right panels illustrate safety attribution hallucination:
over-flagging (assigning an unsupported harmful attribution to benign content),
under-flagging (omitting the harmful attribution from harmful content), and
mis-attribution (assigning the harmfulness to the wrong target despite correctly
identifying the content as harmful). A full taxonomy of hallucination categories
is provided in Section~\ref{sec:metric}.}
    \label{fig:framework}
\end{figure*}

Existing multimodal safety guards typically generate rationales in two ways. One line of work constructs textual rationales through language-based workflows applied to the initial image-text input and its intermediate analyses~\citep{ucot,mind,alarm,prismagent,patmd}. A more recent line of work follows the Think-with-Image paradigm, allowing models to conduct more fine-grained visual inspection through operations or tools such as local cropping, rotation, and OCR before forming the rationale~\citep{visualcot,dyfo,zoomeye,deepeyes,thyme}. Despite these different rationale-generation mechanisms, prior studies primarily evaluate their effectiveness through final classification accuracy, while paying limited attention to hallucination in the generated rationales themselves. As shown in Figure \ref{fig:framework}, such hallucinated rationales may provide an inaccurate explanation for a safety decision and reduce the interpretability and verifiability of multimodal safety guards~\citep{hallucinationsurvey,hallusionbench,faithscore,faithfulcot}.

Motivated by this gap, we first investigate (\textbf{RQ1}) \emph{to what extent do current multimodal safety guards hallucinate rationales when justifying their safety decisions?} To answer this question, we analyze rationale hallucination from two complementary perspectives: visual rationale hallucination and safety attribution hallucination. Specifically, we quantify whether the visual facts described in generated rationales are supported by the input image, and further measure whether the safety-related attributions in rationales are consistent with the ground-truth annotations~\citep{faithscore,haleval}. Our evaluation covers 8 representative rationale-generating safety guard methods, including both recent rationale generation baselines and Think-with-Image approaches, across five vision-language backbones and four safety datasets.

Our results reveal substantial rationale hallucination in current multimodal safety guards. Visual hallucination persists across datasets, methods, and model scales, while higher classification accuracy does not necessarily imply better-grounded rationales. Neither workflow-based nor Think-with-Image methods consistently mitigate this problem. Moreover, even with correctly grounded visual facts, models may still assign unsupported harmful attributions to benign inputs or mis-attribute the harmfulness of correctly classified harmful inputs. These findings show that correct safety predictions do not necessarily imply reliable supporting rationales~\citep{predictiveguardrails,faithfulselfevolvers}.

Having established the prevalence of rationale hallucination, we next investigate (\textbf{RQ2}) \emph{why do current multimodal safety guards hallucinate rationales, and where do these hallucinations arise?} We focus on the Think-with-Image paradigm, which enables models to actively inspect visual information through tool-driven operations and thus provides a representative approach for analyzing the role of visual exploration in rationale generation~\citep{vtsv,deepeyes,pixelreasoner,thyme}. We analyze hallucination from both the rationale content and the rationale-generation process. At the content level, safety-critical attribute and relational facts are
considerably more error-prone than basic visual facts. At the generation level,
Think-with-Image methods use visual tools infrequently and often produce coarse
crops, yielding limited improvements in attribute and relational grounding.  Together, these findings reveal a gap between models' ability to inspect visual information and their ability to generate reliable safety rationales.

To mitigate rationale hallucination, we propose TCG-RL (Tool-Contrastive Grounding Reinforcement Learning), a simple yet effective framework for multimodal safety guards. TCG-RL incorporates hallucination-aware rewards into reinforcement learning and leverages paired tool-use and tool-free rollouts to encourage grounded rationale generation~\citep{grpo,ppo}. Specifically, TCG-RL contrasts rationales generated with and without visual tools, rewarding tool use only when it leads to more grounded rationales. This enables models to selectively invoke visual tools for safety-critical rationale generation. TCG-RL reduces rationale hallucination while maintaining comparable safety prediction accuracy. It improves visual rationale grounding by approximately 20\% over strong baselines and generates more concise and safety-relevant rationales.

In summary, we demonstrate that trustworthy multimodal safety guards require more than accurate predictions: they require rationales that are grounded in the input and support the safety decision. Our systematic analysis reveals substantial visual rationale hallucination and safety attribution hallucination in existing rationale-generating methods, and further exposes the limitations of current Think-with-Image approaches. Building on these findings, TCG-RL learns to selectively acquire visual information and generates more grounded and verifiable rationales, advancing multimodal safety guards from accurate prediction toward trustworthy reasoning.

\section{Rationale Hallucination in Multimodal Safety Guards}
\label{sec:metric}

Given an image-text pair $(I,x)$, a multimodal safety guard generates a
rationale $r$ and a binary safety prediction $\hat{y}\in{0,1}$, where
$1$ denotes harmful content and $0$ denotes benign content. Beyond prediction
accuracy, we study whether the generated rationale contains unsupported
evidence, which may lead to hallucinated explanations of safety decisions. 
Specifically, we characterize evidence hallucination from two
complementary perspectives: \emph{visual evidence hallucination}, which
measures whether the visual facts described in a rationale are supported by the
input image, and \emph{safety attribution hallucination}, which measures
whether the rationale correctly attributes the safety meaning of the observed
content. The former concerns whether a model correctly identifies what appears
in an image, while the latter concerns whether it correctly interprets the
safety implications of those observations. Specific examples are shown in figure \ref{fig:framework}.

\textbf{Visual Evidence Hallucination.}
We first investigate whether the visual evidence described in a rationale is
supported by the input image. Inspired by the verification
process introduced in FaithScore \citep{faithscore}, we decompose each rationale
into atomic visual facts and verify whether these facts can be grounded in the
corresponding image.
Specifically, given a rationale $r$, we first use an LLM to extract descriptive
clauses $d_i$, while excluding subjective judgments and hypothetical statements. Each clause is further
decomposed into atomic visual facts ${(e_i,k_i)}_{i=1}^{n}$, where
$k_i\in{\mathrm{V},\mathrm{A},\mathrm{R}}$ denotes a
\emph{Visual}, \emph{Attribute}, or \emph{Relation} fact, respectively.
A verifier determines whether each fact $e_i$ is supported by image $I$:
$s(e_i,I)\in{0,1}$. We calculate the visual evidence score as:
\begin{equation}
\mathrm{VES}(r,I)=
\frac{1}{n}\sum_{i=1}^{n}s(e_i,I),
\end{equation}
A higher VES score indicates fewer visual evidence hallucination. 

\textbf{Safety Attribution Hallucination.}
Visual grounding alone does not fully capture hallucination in safety
reasoning. A rationale may correctly describe visual contents while providing
an incorrect explanation of why the content is harmful or benign. We define
\emph{safety attribution hallucination} as an unsupported safety-related
attribution in the rationale that is inconsistent with the ground-truth safety
annotation.
Let $\hat{a}(r)$ denote the safety attribution extracted from rationale $r$ by
a judge constrained to the dataset-specific annotation space, and let $a^*$
denote the corresponding ground-truth attribution. If a rationale
explicitly indicates that no harmful attribution is applicable, we set
$\hat{a}(r)=\varnothing$.
We characterize three forms of safety attribution hallucination:
\begin{align}
\mathcal{H}_{\mathrm{OF}}(r,I)
&= \mathbb{I}[y=0 \land \hat{y}=1 \land \hat{a}(r)\neq\varnothing], \\
\mathcal{H}_{\mathrm{UF}}(r,I)
&= \mathbb{I}[y=1 \land \hat{y}=0 \land \hat{a}(r)=\varnothing], \\
\mathcal{H}_{\mathrm{MA}}(r,I)
&= \mathbb{I}[y=1 \land \hat{y}=1 \land \hat{a}(r)\neq a^*].
\end{align}
Here, \emph{over-flagging} ($\mathcal{H}_{\mathrm{OF}}$) refers to assigning
unsupported harmful attributions to benign content; \emph{under-flagging}
($\mathcal{H}_{\mathrm{UF}}$) refers to failing to identify supported harmful
attributions; and \emph{mis-attribution} ($\mathcal{H}_{\mathrm{MA}}$) refers
to assigning an incorrect harmful attribution despite a correct harmful
prediction.
We compute each hallucination type as its empirical rate over the corresponding
subset, denoted as $\overline{\mathcal{H}}$. The overall safety attribution
score is defined as:
\begin{equation}
\mathrm{SAS}
=1-\frac{1}{3}
\left(
\overline{\mathcal{H}}_{\mathrm{OF}}+
\overline{\mathcal{H}}_{\mathrm{UF}}+
\overline{\mathcal{H}}_{\mathrm{MA}}
\right).
\end{equation}
A higher SAS indicates fewer safety attribution hallucinations.

\section{Evaluation of Rationale Hallucination}
\label{evaluation}

We first conduct a systematic empirical study to answer \textbf{RQ1}: \emph{To what extent do current multimodal safety auditors hallucinate evidence when justifying their safety decisions?}

\subsection{Experimental Setup}

\textbf{Benchmarks.}
We evaluate rationale-augmented binary safety classification and rationale hallucination on four image-text multimodal safety benchmarks: Harm-C and Harm-P from HarMeme \citep{harmeme}, MAMI \citep{mami}, and PrideMM \citep{pridemm}. Further dataset details are provided in Appendix \ref{app:datasets}.

\textbf{Baseline Methods.}
We evaluate three types of rationale-generation methods. First, \emph{Direct CoT} directly generates a textual rationale from the image-text input before making the safety prediction. Second, we consider three workflow-based methods that enhance rationale generation through additional reasoning or adaptation procedures: U-CoT+ \citep{ucot}, MIND \citep{mind}, and ALARM \citep{alarm}. Finally, we evaluate five Think-with-Image methods that actively inspect the image through iterative visual exploration or tool use: Visual CoT \citep{visualcot}, DyFo \citep{dyfo}, ZoomEye \citep{zoomeye}, DeepEyes \citep{deepeyes}, and Thyme \citep{thyme}. For all methods, we retain their original reasoning and tool-use procedures. Further baseline details are provided in Appendix \ref{app:baselines}.

\textbf{Implementation Details.}
For fair comparison, we standardize the final prediction and rationale format across methods.
Large-scale open-weight models are accessed through hosted APIs, while other open-weight models are
deployed locally with vLLM \citep{vllm} on NVIDIA H20 GPUs.
Additional implementation details are provided in Appendix \ref{app:implementation}.

\subsection{Visual Evidence Hallucination}

We first examine whether the visual evidence described in safety rationales is supported by the input image. Table~\ref{tab:evaluation_results} reports the results with Qwen2.5-VL-7B and Qwen2.5-VL-32B \citep{qwen25}, covering the Direct CoT baseline, recent Rationale Generation methods, and Think-with-Image methods. Results for additional model backbones and settings are provided in Appendix~\ref{app:additional_results_visual}.

\textbf{Visual hallucination is prevalent in generated rationales.}
Across datasets, methods, and model scales, generated rationales consistently contain visual facts that cannot be verified from the input image. This pattern persists across different safety scenarios and rationale-generation strategies, showing that seemingly plausible rationales may still hallucinate substantial visual information.


\textbf{Higher classification accuracy does not necessarily correspond to better rationale grounding.} Across many method--dataset combinations, gains in classification accuracy are accompanied by unchanged or lower VES. This discrepancy is observed for both workflow-based rationale generation and Think-with-Image methods, indicating that better safety prediction can be achieved without better grounding of the generated rationale. Therefore, classification accuracy and rationale grounding capture distinct properties of multimodal safety guards, and prediction performance alone cannot determine the reliability of the generated rationale.


\textbf{Existing rationale-generation methods do not consistently mitigate visual hallucination.} Compared with Direct CoT, workflow-based methods frequently improve classification accuracy without improving VES. Think-with-Image methods, despite explicitly introducing visual inspection, also fail to consistently achieve higher VES. While some methods improve visual grounding in individual settings, these gains do not generalize across datasets or backbones. These results suggest that more sophisticated reasoning workflows or access to visual tools alone is insufficient to reliably reduce visual hallucination. We further investigate the underlying causes in Section~\ref{cause}.


\textbf{Scaling alone does not resolve rationale hallucination.}
Moving from Qwen2.5-VL-7B to Qwen2.5-VL-32B does not lead to systematic
improvements in VES across methods and datasets, and substantial visual
hallucination remains with the larger backbone. This indicates that increasing model capacity alone does not reliably improve the grounding of generated rationales, making rationale hallucination a persistent issue across model scales.



\begin{safetakeaways}
  \textbf{\textit{Takeaways:}} \textit{Visual hallucination persists across
  datasets, rationale-generation methods, and model scales. Since improved
  safety accuracy does not reliably imply better grounding, rationale
  faithfulness is a distinct reliability dimension beyond classification
  accuracy.}
\end{safetakeaways}

\begin{table*}[t]
\caption{Classification Accuracy (Acc.), Visual Evidence Score (VES), and Safety Attribution Score (SAS) across image-text safety benchmarks. Higher values are better. Values in parentheses denote changes relative to Direct CoT.}
\label{tab:evaluation_results}
\centering
\setlength{\extrarowheight}{0pt}
\resizebox{\linewidth}{!}{
\begin{tabular}{lccc|ccc|ccc|ccc}
\toprule
&
\multicolumn{3}{c|}{\textbf{Harm-C}} &
\multicolumn{3}{c|}{\textbf{Harm-P}} &
\multicolumn{3}{c|}{\textbf{MAMI}} &
\multicolumn{3}{c}{\textbf{PrideMM}} \\
\cmidrule(lr){2-4}
\cmidrule(lr){5-7}
\cmidrule(lr){8-10}
\cmidrule(lr){11-13}
&
\textbf{Acc.} & \textbf{VES} & \textbf{SAS} &
\textbf{Acc.} & \textbf{VES} & \textbf{SAS} &
\textbf{Acc.} & \textbf{VES} & \textbf{SAS} &
\textbf{Acc.} & \textbf{VES} & \textbf{SAS} \\
\midrule

\multicolumn{13}{l}{\textbf{Qwen2.5-VL-7B}} \\

\texttt{Direct CoT}
&66.4 &73.3 &80.7
&56.9 &76.4 &76.6
&71.4 &\textbf{81.4} &75.9
&60.2 &75.7 &72.7 \\

\multicolumn{13}{l}{\emph{Workflow-based Baselines}} \\

\texttt{U-CoT+}
&\cellup{72.0}{5.6} &\celldown{63.9}{9.4} &\cellup{84.8}{4.1}
&\cellup{57.1}{0.2} &\celldown{68.4}{8.0} &\celldown{75.9}{0.7}
&\cellzero{71.4} &\celldown{77.9}{3.5} &\cellup{76.9}{1.0}
&\cellup{60.9}{0.7} &\celldown{68.5}{7.2} &\celldown{63.0}{9.7} \\

\texttt{MIND}
&\cellup{68.4}{2.0} &\celldown{68.3}{5.0} &\celldown{75.0}{5.7}
&\cellup{57.2}{0.3} &\cellup{78.4}{2.0} &\celldown{71.8}{4.8}
&\cellup{72.7}{1.3} &\celldown{80.3}{1.1} &\celldown{74.3}{1.6}
&\cellup{64.7}{4.5} &\celldown{74.5}{1.2} &\celldown{62.0}{10.7} \\

\texttt{ALARM}
&\cellzero{66.4} &\celldown{73.0}{0.3} &\celldown{75.2}{5.5}
&\cellup{60.3}{3.4} &\celldown{74.6}{1.8} &\celldown{76.0}{0.6}
&\cellup{72.4}{1.0} &\celldown{79.1}{2.3} &\celldown{75.1}{0.8}
&\cellup{64.3}{4.1} &\celldown{70.8}{4.9} &\celldown{66.4}{6.3} \\

\multicolumn{13}{l}{\emph{Think-with-Image Paradigm}} \\

\texttt{Visual CoT}
&\cellup{70.3}{3.9} &\cellup{74.5}{1.2} &\celldown{76.0}{4.7}
&\cellup{59.2}{2.3} &\celldown{75.6}{0.8} &\celldown{73.6}{3.0}
&\celldown{69.4}{2.0} &\celldown{81.1}{0.3} &\celldown{69.1}{6.8}
&\cellzero{60.2} &\celldown{75.6}{0.1} &\celldown{71.4}{1.3} \\

\texttt{DyFo}
&\cellup{67.8}{1.4} &\celldown{71.2}{2.1} &\celldown{80.6}{0.1}
&\cellzero{56.9} &\celldown{74.2}{2.2} &\cellup{78.1}{1.5}
&\cellup{72.0}{0.6} &\celldown{79.4}{2.0} &\celldown{71.0}{4.9}
&\cellup{63.1}{2.9} &\celldown{74.5}{1.2} &\celldown{71.9}{0.8} \\

\texttt{ZoomEye}
&\celldown{61.3}{5.1} &\celldown{70.3}{3.0} &\celldown{80.5}{0.2}
&\cellup{57.5}{0.6} &\celldown{72.9}{3.5} &\celldown{76.2}{0.4}
&\cellup{71.7}{0.3} &\celldown{75.0}{6.4} &\celldown{75.2}{0.7}
&\cellup{60.9}{0.7} &\celldown{65.5}{10.2} &\celldown{69.1}{3.6} \\

\texttt{Thyme}
&\cellup{\textbf{78.9}}{12.5} &\celldown{71.4}{1.9} &\celldown{79.9}{0.8}
&\cellup{\textbf{65.1}}{8.2} &\celldown{76.2}{0.2} &\celldown{72.0}{4.6}
&\cellup{\textbf{78.3}}{6.9} &\celldown{80.7}{0.7} &\celldown{70.7}{5.2}
&\cellup{\textbf{74.6}}{14.4} &\cellup{77.5}{1.8} &\celldown{65.3}{7.4} \\

\midrule

\multicolumn{13}{l}{\textbf{Qwen2.5-VL-32B}} \\

\texttt{Direct CoT}
&60.5 &69.1 &87.5
&60.3 &74.5 &85.3
&69.3 &\textbf{80.6} &89.4
&62.5 &73.8 &80.5 \\

\multicolumn{13}{l}{\emph{Workflow-based Baselines}} \\

\texttt{U-CoT+}
&\cellup{72.6}{12.1} &\celldown{59.1}{10.0} &\celldown{87.3}{0.2}
&\cellup{60.6}{0.3} &\celldown{65.3}{9.2} &\celldown{80.3}{5.0}
&\cellup{73.3}{4.0} &\celldown{68.8}{11.8} &\celldown{84.1}{5.3}
&\cellup{66.7}{4.2} &\celldown{63.9}{9.9} &\cellup{80.6}{0.1} \\

\texttt{MIND}
&\cellup{68.6}{8.1} &\celldown{55.6}{13.5} &\celldown{82.9}{4.6}
&\cellup{60.8}{0.5} &\celldown{60.2}{14.3} &\celldown{81.7}{3.6}
&\cellup{76.0}{6.7} &\celldown{69.3}{11.3} &\celldown{81.7}{7.7}
&\cellup{70.8}{8.3} &\celldown{69.1}{4.7} &\celldown{78.2}{2.3} \\

\texttt{ALARM}
&\cellup{66.4}{5.9} &\cellup{73.3}{4.2} &\celldown{74.0}{13.5}
&\cellzero{60.3} &\cellup{74.8}{0.3} &\celldown{82.8}{2.5}
&\cellup{72.4}{3.1} &\celldown{78.6}{2.0} &\celldown{74.0}{15.4}
&\cellup{64.3}{1.8} &\celldown{71.4}{2.4} &\celldown{67.1}{13.4} \\

\multicolumn{13}{l}{\emph{Think-with-Image Paradigm}} \\

\texttt{Visual CoT}
&\cellup{63.8}{3.3} &\cellup{70.4}{1.3} &\celldown{86.1}{1.4}
&\cellup{62.0}{1.7} &\cellup{75.5}{1.0} &\celldown{84.6}{0.7}
&\cellup{73.0}{3.7} &\celldown{79.2}{1.4} &\celldown{88.1}{1.3}
&\cellup{63.9}{1.4} &\celldown{72.2}{1.6} &\celldown{79.1}{1.4} \\

\texttt{DyFo}
&\cellup{63.3}{2.8} &\celldown{67.2}{1.9} &\celldown{87.4}{0.1}
&\cellup{63.9}{3.6} &\celldown{72.5}{2.0} &\cellup{86.5}{1.2}
&\cellup{74.8}{5.5} &\celldown{78.8}{1.8} &\celldown{86.3}{3.1}
&\cellzero{62.5} &\celldown{72.0}{1.8} &\celldown{78.7}{1.8} \\

\texttt{ZoomEye}
&\cellup{61.3}{0.8} &\celldown{56.8}{12.3} &\celldown{83.8}{3.7}
&\celldown{59.4}{0.9} &\celldown{59.0}{15.5} &\celldown{84.1}{1.2}
&\cellup{69.5}{0.2} &\celldown{65.3}{15.3} &\celldown{88.2}{1.2}
&\cellup{63.5}{1.0} &\celldown{61.9}{11.9} &\celldown{80.2}{0.3} \\

\texttt{DeepEyes}
&\cellup{\textbf{75.4}}{14.9} &\celldown{66.0}{3.1} &\celldown{80.3}{7.2}
&\cellup{\textbf{64.2}}{3.9} &\celldown{68.3}{6.2} &\celldown{78.3}{7.0}
&\cellup{\textbf{82.8}}{13.5} &\celldown{74.1}{6.5} &\celldown{79.1}{10.3}
&\cellup{\textbf{72.0}}{9.5} &\celldown{70.9}{2.9} &\celldown{71.7}{8.8} \\

\bottomrule
\end{tabular}
}
\end{table*}


\subsection{Safety Attribution Hallucination}


Visual grounding captures only whether the visual facts in a rationale are supported by the input image; it does not determine whether these facts are correctly interpreted from a safety perspective. We therefore further examine \emph{safety attribution hallucination}, asking whether a rationale can still misinterpret why an input is harmful or benign even when its visual description is grounded. Following Section~\ref{sec:metric}, we evaluate three forms of safety attribution hallucination: \emph{over-flagging}, \emph{under-flagging}, and \emph{mis-attribution}. Importantly, this evaluation is conducted on rationales \emph{without} visual hallucination, allowing us to isolate errors in safety attribution from errors in visual perception. Table ~\ref{tab:evaluation_results} and Figure~\ref{fig:safety_attribution} reports the corresponding results, with additional models provided in Appendix~\ref{app:additional_results_safety}.

\textbf{Safety attribution hallucination remains prevalent even when visual facts are correctly grounded.}
Across datasets, methods, and model scales, safety attribution scores remain far from perfect, revealing a distinct source of rationale hallucination beyond incorrect visual descriptions. Neither workflow-based rationale generation nor Think-with-Image methods consistently improves safety attribution over \texttt{Direct CoT}, and higher classification accuracy is frequently accompanied by unchanged or degraded attribution scores. These results show that correctly grounding what appears in an image does not necessarily ensure a correct interpretation of its safety implications.


\textbf{Models are more prone to fabricating harmfulness than overlooking it.}
\emph{Over-flagging} scores are generally lower than \emph{under-flagging} scores, indicating that models are more likely to assign unsupported harmful attributions to benign inputs than to overlook harmful attributions in harmful inputs. Such errors include fabricating harmful intent or exaggerating the safety implications of otherwise benign content. Detailed cases are provided in Appendix~\ref{app:case_study}.


\textbf{Correct harmful predictions can still be supported by incorrect
safety attributions.} \emph{Mis-attribution} remains substantial across several datasets and methods: a model may correctly classify an input as harmful while attributing its harmfulness to the wrong target, category, or stance. In such cases, the final safety label is correct, but the rationale explains the decision for the wrong reason. This further demonstrates that classification correctness and safety attribution are distinct properties. Representative cases are provided in Appendix~\ref{app:case_study}.



\begin{safetakeaways}
  \textbf{\textit{Takeaways:}} \textit{Visual-evidence and safety-attribution hallucinations are distinct but complementary: models may misidentify \emph{what} is present or \emph{why} it is harmful. Thus, even a correct safety prediction does not guarantee a reliable rationale.}
\end{safetakeaways}

\begin{figure*}[t]
    \centering
    \includegraphics[width=\textwidth]{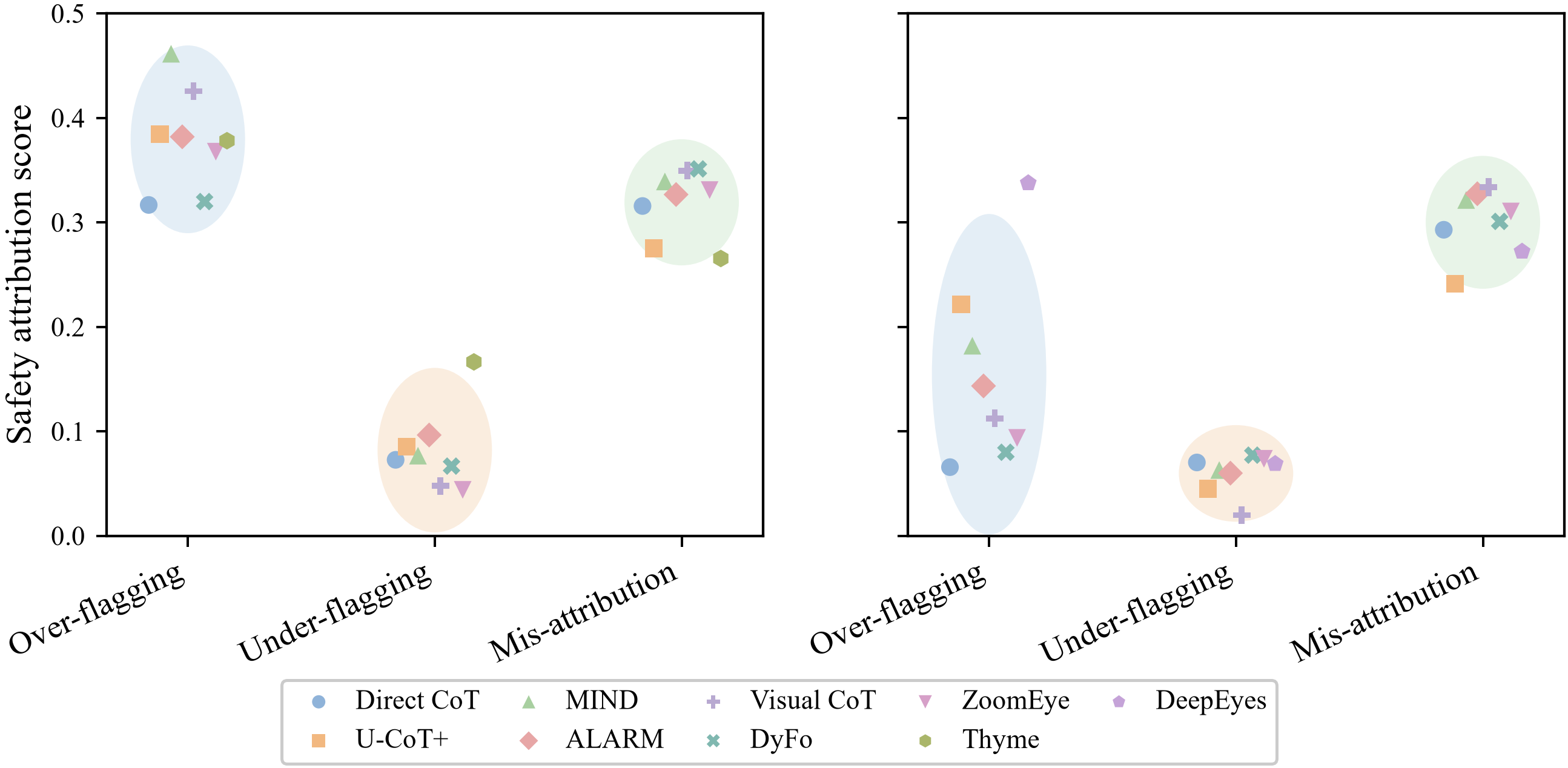}
    \caption{Mean safety-attribution scores across the four image-text safety benchmarks for Qwen2.5-VL-7B (left) and Qwen2.5-VL-32B (right). Each point represents a rationale-generation method, and
the shaded ellipses highlight the distribution within each category. Lower is better.}
    \label{fig:safety_attribution}
\end{figure*}


\section{Understanding the Causes of Rationale Hallucination}
\label{cause}

The results in Section~\ref{evaluation} show that rationale hallucination persists across multimodal safety guards, even when more sophisticated rationale-generation or visual-inspection mechanisms are introduced. We next turn to (\textbf{RQ2}): \emph{why do current multimodal safety guards hallucinate rationales, and where do these hallucinations arise?} We focus on the Think-with-Image paradigm and examine the problem from two complementary perspectives: \emph{what types of visual facts are most prone to hallucination}, and \emph{how models acquire visual information through tool use during rationale generation}.


\subsection{Hallucination across Atomic Fact Categories}
\label{sec:category_analysis}





\begin{wrapfigure}[23]{r}{0.5\textwidth}
  \centering
  \includegraphics[width=\linewidth]{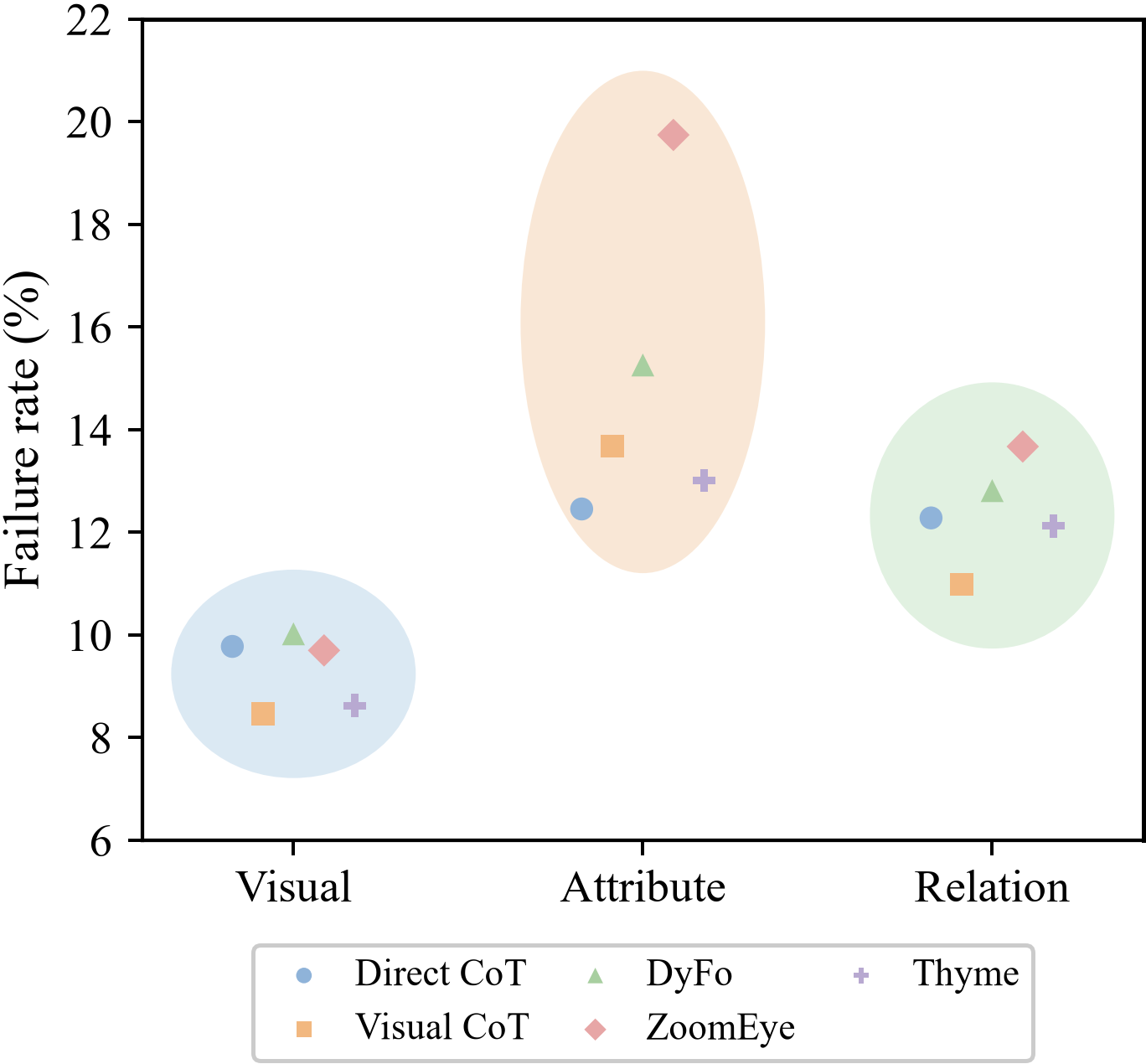}
  \caption{Atomic-fact failure rates on Harm-P with Qwen2.5-VL-7B. Each marker denotes a rationale-generation method, and
the shaded ellipses highlight the distribution within each category. Lower is better.
}
  \label{fig:hallucination_category}
\end{wrapfigure}

We first examine where visual hallucination is concentrated.
Figure~\ref{fig:hallucination_category} reports the failure rates of three
atomic-fact categories: \emph{Visual}, \emph{Attribute}, and \emph{Relation}. Across datasets and methods, Attribute and Relation facts consistently show higher failure rates than Visual facts. Visual facts mainly capture the presence or number of entities, whereas Attribute and Relation facts require finer-grained understanding of object properties, interactions, and spatial or semantic relationships. Importantly, 
these finer-grained facts often determine how the visual content should be interpreted from a safety perspective: an incorrect attribute or relation may lead to an incorrect target, stance, or harm 
attribution. Thus, rationale hallucination is particularly concentrated in the types of visual information that are most relevant to safety interpretation.

Think-with-Image methods provide only limited improvement on these
fine-grained facts. Although visual inspection can reduce failures on Visual
facts in some settings, the improvement does not consistently extend to
Attribute and Relation facts, which remain highly error-prone and sometimes
even deteriorate relative to \texttt{Direct CoT}. This suggests that enabling a model to inspect the image more closely does not by itself ensure that it can correctly resolve the properties and relationships required for safety reasoning. This observation motivates us to further examine how visual tools are actually used during rationale generation.

\subsection{Tool Usage in Think-with-Image Safety Guards}
\label{sec:tool_analysis}

\begin{table*}[t]
\caption{
Tool-use statistics of \texttt{DyFo} across model backbones and safety datasets.
Higher values indicate more frequent tool use or more focused cropping.
}
\label{tab:dyfo_tool_usage}
\centering
\small
\setlength{\tabcolsep}{5pt}
\resizebox{\linewidth}{!}{
\begin{tabular}{lcc|cc|cc}
\toprule
\multirow{2}{*}{\textbf{Backbone}}
& \multicolumn{2}{c|}{\textbf{Harm-P}}
& \multicolumn{2}{c|}{\textbf{MAMI}}
& \multicolumn{2}{c}{\textbf{PrideMM}} \\
\cmidrule(lr){2-3}
\cmidrule(lr){4-5}
\cmidrule(lr){6-7}
& Tool Use & Focused Zoom
& Tool Use & Focused Zoom
& Tool Use & Focused Zoom \\
\midrule
\texttt{Qwen2.5-VL-7B}
& 21.2 & 11.9
& 28.0 & 13.2
& 26.0 & 15.0 \\

\texttt{Qwen2.5-VL-32B}
& 34.8 & 16.4
& 42.7 & 18.0
& 42.5 & 22.1 \\

\texttt{Qwen3-8B-VL}
& 34.5 & 18.1
& 49.6 & 24.8
& 49.5 & 23.8 \\

\texttt{Qwen3-32B-VL}
& 51.3 & 27.6
& 65.4 & 40.5
& 66.1 & 40.2 \\

\texttt{Qwen3.5-27B}
& 4.6 & 3.4
& 1.6 & 1.5
& 13.0 & 10.3 \\
\bottomrule
\end{tabular}
}
\end{table*}

We next investigate whether the limited gains of Think-with-Image methods can be traced to their tool-use behavior. We focus on \texttt{DyFo}~\citep{dyfo} because it allows the model to decide autonomously whether and how to invoke visual tools, including operations such as cropping, rotation, and OCR. We report the tool-use rate as the proportion of samples in which the model invokes at 
least one visual tool. To characterize how precisely the model localizes visual information,  we additionally report the focused-zoom rate, defined as the proportion of samples whose selected crop covers less than $85\%$ of the original image.
As shown in Table~\ref{tab:dyfo_tool_usage}, tool use is both infrequent and inconsistent across backbones and datasets: models often generate rationales without additional visual inspection, and do not adopt a consistent
inspection policy despite having access to the same tools.

Even when tools are invoked, visual inspection is often insufficiently focused. Focused-zoom rates remain considerably lower than overall tool-use
rates, showing that many crop operations retain a large portion of the original image rather than isolating a specific safety-relevant region. As a result, invoking a tool does not necessarily provide the fine-grained visual information needed to resolve difficult attributes or relations. More broadly, tool availability alone is insufficient: models must also determine \emph{when} additional inspection is needed, \emph{where} to inspect, and \emph{how} to incorporate the resulting observation into the rationale.

In summary, our analysis identifies two closely related bottlenecks behind rationale hallucination. At the content level, hallucination is concentrated in fine-grained attributes and relations that are critical for safety interpretation. At the acquisition level, current Think-with-Image methods do not consistently invoke visual tools or focus them on informative regions. Together, these findings suggest that reducing rationale hallucination requires more than providing visual tools---it requires selectively acquiring the visual information that is actually needed for safety reasoning.

\section{Tool-Contrastive Grounding Reinforcement Learning}
\label{sec:tcg_rl}


The analyses in Sections~\ref{evaluation} and~\ref{cause} reveal that
providing visual tools alone does not reliably reduce rationale hallucination. Current Think-with-Image safety guards may fail to invoke tools when additional inspection is needed, and tool use itself does not necessarily lead to better visual grounding. This suggests that the key challenge is not simply to increase tool usage, but to encourage tool use only when it contributes meaningful grounding information. To this end, we propose \emph{Tool-Contrastive Grounding Reinforcement Learning} (TCG-RL), which explicitly rewards grounding improvements brought by visual tool use while preserving the task-level safety objective.

\subsection{Tool-Contrastive Grounding Reward}
\label{sec:tcg_reward}

Given an image-text pair $(I,x)$, the policy $\pi_\theta$ generates a trajectory $\tau$. TCG-RL retains the tool-augmented reasoning process of Think-with-Image paradigm, the model may iteratively call image operations such as cropping, resizing, and rotation before producing its final answer. We measure the visual grounding of a trajectory using the atomic-fact faithfulness score defined in Section~\ref{sec:metric}, and denote this per-trajectory score by
$F_{\mathrm{atomic}}(\tau,I)$. The contrastive reward uses only
the atomic visual-fact score, which ensures that the contrastive signal
is computed independently for each sample and is not affected by batch-level fluctuations in attribution judgments.

To determine whether tool use provides a meaningful grounding benefit, we
construct an reference score for each training example. Specifically,
a fixed no-tool model is prompted with the same image-text pair
using a direct-answer prompt and generates a reference rationale without
executing visual tools. The reference rationale is evaluated with the same
atomic-fact faithfulness procedure, yielding
$F_{\mathrm{ref}}(q)$ for question identifier $q$. 
For a trajectory $\tau$ associated with question $q$, we define the relative
grounding improvement and the corresponding contrastive reward as:
\begin{equation}
\Delta(\tau,q)=F_{\mathrm{atomic}}(\tau,I)-F_{\mathrm{ref}}(q),
\qquad
R_{\mathrm{TCG}}(\tau,q)
=\lambda_{\mathrm{cf}}\,
\begin{cases}
+1, & \Delta(\tau,q)>\epsilon,\\
0, & |\Delta(\tau,q)|\leq\epsilon,\\
-1, & \Delta(\tau,q)<-\epsilon ,
\end{cases}
\label{eq:tcg_reward}
\end{equation}
where $\epsilon$ defines a tolerance margin for small or potentially noisy differences, and $\lambda_{\mathrm{cf}}$ controls the strength of the
contrastive signal. In this way, TCG-RL rewards a trajectory only when visual tool use yields a sufficiently large grounding improvement over its no-tool reference, while penalizing tool use that degrades grounding. The objective is therefore selective rather than frequent tool use.

We combine this contrastive grounding signal with the standard task-level rewards. Let $r_{\mathrm{fmt}}(\tau)$ denote the format reward,
$r_{\mathrm{vqa}}(\tau)$ the safety-prediction reward, and $F(\tau)$ the
trajectory-level grounding reward. The total reward is defined as:
\begin{equation}
R(\tau)
=
\lambda_{\mathrm{fmt}}r_{\mathrm{fmt}}(\tau)
+\lambda_{\mathrm{vqa}}r_{\mathrm{vqa}}(\tau)
+\lambda_{\mathrm{faith}}F(\tau)
+R_{\mathrm{TCG}}(\tau,q).
\label{eq:tcg_total_reward}
\end{equation}
We optimize the policy using Group Relative Policy Optimization
(GRPO) \citep{grpo}.
For each image-text input $(I_i,x_i)$, the policy samples a group of
$G$ trajectories $\{\tau_{i,1},\ldots,\tau_{i,G}\}$.
Each trajectory receives the total reward defined in
Eq.~\ref{eq:tcg_total_reward}.
The implementation details of reference construction,
hallucination judging, and hyperparameters are provided in
Appendix~\ref{app:implementation_tcg}.

\subsection{Experimental Results on TCG-RL}
\label{sec:tcg_results}

\textbf{Experimental setup.}
We evaluate TCG-RL on four image-text safety benchmarks: Harm-C, Harm-P,
MAMI, and PrideMM, using Qwen2.5-VL-7B as the backbone. We compare against
Direct CoT, standard GRPO, and Thyme, a representative Think-with-Image
training method. We additionally include TCG-RL without the tool-contrastive
reward (\textsc{TCG-RL w/o TC}) to isolate the contribution of the proposed
reward. All reinforcement learning experiments are conducted on eight NVIDIA
H20 96GB GPUs. In Table \ref{tab:tcg_results}, we report classification Accuracy, Visual Evidence Score (VES), and Safety Attribution Score (SAS), with higher values indicating better
performance for all metrics.

\begin{table*}[t]
\centering
\caption{
Comparison of TCG-RL with Direct CoT, standard GRPO,
Thyme, and the variant without the tool-contrastive reward (TCG-RL w/o TC).
The best and second-best results in each column are highlighted in
\textbf{bold} and \underline{underlined}, respectively.
}
\label{tab:tcg_results}
\resizebox{\textwidth}{!}{
\begin{tabular}{lccc|ccc|ccc|ccc}
\toprule
\multirow{2}{*}{\textbf{Method}}
& \multicolumn{3}{c|}{\textbf{Harm-C}}
& \multicolumn{3}{c|}{\textbf{Harm-P}}
& \multicolumn{3}{c|}{\textbf{MAMI}}
& \multicolumn{3}{c}{\textbf{PrideMM}} \\
\cmidrule(lr){2-4}
\cmidrule(lr){5-7}
\cmidrule(lr){8-10}
\cmidrule(lr){11-13}
& Acc. & VES & SAS
& Acc. & VES & SAS
& Acc. & VES & SAS
& Acc. & VES & SAS \\
\midrule
\texttt{Direct CoT}
& 66.4 & 73.3 & 80.7
& 56.9 & 76.4 & \underline{76.6}
& 71.4 & 81.4 & 75.9
& 60.2 & 75.7 & 72.7 \\

\texttt{GRPO}
& 74.0 & 71.2 & 80.6
& 61.4 & 74.8 & 69.4
& 76.6 & 79.9 & 75.9
& 69.6 & 75.6 & 72.4 \\

\texttt{Thyme}
& \textbf{78.9} & 71.4 & 79.9
& \underline{65.1} & 76.2 & 72.0
& \textbf{78.3} & 80.7 & 70.7
& \underline{74.6} & 77.5 & 65.3 \\

\texttt{TCG-RL w/o TC}
& 74.8 & \underline{81.9} & \underline{82.0}
& 64.5 & \underline{88.1} & 75.3
& \underline{78.2} & \underline{90.0} & \textbf{87.4}
& 71.8 & \underline{86.7} & \underline{75.0} \\

\textbf{\texttt{TCG-RL}}
& \underline{78.0} & \textbf{94.5} & \textbf{82.7}
& \textbf{65.3} & \textbf{97.0} & \textbf{76.8}
& 77.3 & \textbf{97.8} & \underline{86.0}
& \textbf{76.8} & \textbf{95.8} & \textbf{78.5} \\
\bottomrule
\end{tabular}
}
\end{table*}

\textbf{TCG-RL substantially improves rationale reliability across all four benchmarks.}
It consistently achieves the best VES across datasets while maintaining strong SAS, showing that TCG-RL can substantially reduce visual hallucination without introducing a corresponding degradation in safety attribution. These results demonstrate that TCG-RL improves both the grounding and overall reliability of generated rationales across different safety scenarios. Qualitative examples of this improvement are provided in Appendix~\ref{app:case_tcg}.

\textbf{The tool-contrastive reward is critical to the grounding improvement of TCG-RL.}
Removing this component consistently degrades VES across all benchmarks,
showing that the gains cannot be explained by reinforcement learning with
hallucination-aware rewards alone. By explicitly comparing tool-augmented
trajectories against their no-tool references, the proposed reward provides a direct learning signal for identifying when visual tool use actually improves rationale grounding.

\textbf{Improved rationale grounding does not come at the cost of safety
prediction accuracy.} TCG-RL maintains competitive classification performance across all benchmarks and achieves the best Accuracy in several settings, indicating that its gains in rationale grounding are achieved without sacrificing safety prediction performance. 

\section{Related Works}

\textbf{Multimodal Safety Guards and Rationale Generation.}
Early multimodal safety guards primarily formulate safety detection as a
classification problem, predicting whether an image-text pair is harmful or
benign~\citep{harmeme,mami,pridemm,adasteer}. More recent work has extended this setting by introducing explicit
rationale generation to support safety decisions
\citep{memeintel,mememind,expohm}.
For example, U-CoT+ \citep{ucot}, MIND \citep{mind}, and ALARM
\citep{alarm} introduce additional reasoning, retrieval, multi-agent, or
self-improvement workflows to enhance safety prediction and rationale
generation\citep{selfevolvesafety}.
These studies demonstrate the potential of explicit rationales for
multimodal safety guarding, but primarily evaluate their effectiveness through final classification performance. Whether the generated rationales themselves contain hallucinated visual facts or incorrect safety attributions remains largely unexplored.

\textbf{Multimodal Hallucination Evaluation.}
Multimodal hallucination evaluation has studied whether generated text is
supported by visual inputs \citep{hallucinationsurvey}. Closed-vocabulary
benchmarks and metrics~\citep{chair,amber,hallusionbench} assess errors involving object existence, attributes, and relations. For open-ended generation, FaithScore
\citep{clipscore,faithscore} decomposes a response into atomic facts and
verifies each fact against the image. Other work further considers event-level hallucination \citep{haleval,mememind,muphi}.
These methods provide effective tools for identifying hallucinated visual content, but do not capture a safety-specific failure: a rationale may accurately describe the visual content while still assigning an incorrect safety attribution~\citep{personalizationrisk}. Our work extends hallucination analysis from visual facts to the safety interpretation built upon them.

\textbf{Think-with-Image and Tool-Augmented Visual Reasoning.}
Think-with-Image methods augment visual reasoning by allowing models to
iteratively inspect images rather than relying solely on a holistic visual
representation. Visual CoT \citep{visualcot} and DyFo \citep{dyfo}, for
example, use progressive visual inspection and cropping to resolve fine-grained details. Recent approaches further integrate visual tool use with planning, verification, or post-training \citep{vtsv,deepeyes,pixelreasoner,thyme,zoomeye}.
Although these methods improve performance on general visual reasoning
tasks, existing training objectives do not explicitly distinguish visual tool use that improves rationale grounding from tool use that is redundant. TCG-RL addresses this limitation by contrasting tool-augmented
trajectories with their no-tool references and explicitly rewarding grounding improvements introduced by visual inspection.

\section{Conclusion}

In this work, we systematically investigate rationale hallucination in multimodal safety guards and show that correct safety predictions do not necessarily imply reliable rationales. We characterize this problem through visual rationale hallucination and safety attribution hallucination, and further reveal that current Think-with-Image methods remain limited by insufficient and poorly focused visual exploration. Based on these findings, we propose TCG-RL, which encourages selective visual tool use by rewarding grounding improvements over no-tool references. Experiments across multiple safety benchmarks demonstrate that TCG-RL substantially reduces rationale hallucination while preserving strong safety prediction performance. Overall, our study highlights rationale reliability as an important dimension of multimodal safety guarding beyond classification accuracy.

\subsection*{AI use statement}

In this work, we used generative AI tools to assist with implementing the
proposed methods, translating and polishing technical text, and interpreting
experimental results. We did not use generative AI tools to develop theoretical
models or conceptual frameworks, propose or refine hypotheses, design or
provide feedback on the research methodology or experiments, generate or clean
datasets, or conduct qualitative or thematic data analysis. We did not use
generative AI tools to generate synthetic datasets, formulate mathematical
claims, provide essential ingredients for mathematical proofs, or assist in
writing proofs; these tasks were not applicable to this work.

Additionally, we used generative AI tools to suggest experimental parameters,
create and edit software code, summarize and analyze related literature,
identify research topics and gaps, brainstorm ideas, search for relevant
information, format references, suggest the structure of the paper, improve
the title and keywords, and edit the manuscript for readability. AI assistance
was also used to draft and refine portions of the manuscript, including the
abstract, conclusion, figure captions, and qualitative case-study descriptions.
All AI-assisted code, analyses, interpretations, and written material were
reviewed and verified by the authors. Experimental results were independently
checked against the implemented methods and reported measurements, and all
claims and case-study descriptions were revised where necessary. The authors
take full responsibility for the final content of this work, including all
text, claims, code, analyses, figures, and other artifacts produced with the
aid of generative AI.

\subsection*{Ethics statement}

\paragraph{Societal benefits.}
This work aims to improve the reliability and transparency of multimodal safety
guards. By evaluating both the visual grounding of generated rationales and the
correctness of their safety attributions, our study can help identify failure
modes that may be missed by classification accuracy alone. The proposed
TCG-RL framework further encourages models to use visual tools selectively and
to produce more faithful explanations, which may support safer content
moderation and reduce the burden on human reviewers.

\paragraph{Potential risks and misuse.}
Multimodal safety guards may be misused for automated censorship, surveillance,
or high-stakes moderation without sufficient human oversight. In addition,
hallucinated rationales may make an incorrect decision appear well justified,
thereby increasing the risk of unjustified content removal or refusal. Our
methods are intended for research on safety evaluation and assistance, not for
fully autonomous moderation or decisions affecting individuals without human
review.

\paragraph{Bias and deployment considerations.}
The datasets used in this work contain harmful memes involving political
figures, misogyny, public-health topics, and LGBTQ+-related content. Such data
may reflect cultural, demographic, and annotator biases, and the corresponding
labels and fine-grained attributions may not be universally applicable across
different communities or contexts. Safety judgments are also dependent on the
underlying annotation guidelines and may vary across languages, cultures, and
platform policies. Therefore, model outputs should be interpreted as
policy- and dataset-dependent signals rather than objective judgments of
harmfulness. Responsible deployment requires human oversight, regular
evaluation, transparent moderation policies, and calibration to the target
community.

\paragraph{Use of datasets.}
We use the HarMeme, MAMI, and PrideMM benchmarks, including the Harm-C and
Harm-P subsets of HarMeme, strictly for research and evaluation of multimodal
safety guarding. We follow the terms of use and licensing conditions specified
by the respective dataset creators. The data are used to study harmful-content
detection and rationale reliability, and not to promote, generate, or amplify
harmful content.

\paragraph{Research integrity.}
The examples and analyses reported in this paper are used to characterize
failure modes of multimodal safety guards. We do not endorse the harmful,
derogatory, or hateful content contained in the benchmark data. We take
responsibility for the ethical use of the datasets, the accuracy of the
reported analyses, and the responsible presentation of potentially harmful
examples.

\subsection*{Reproducibility statement}

Detailed information on dataset usage is provided in Appendix \ref{app:datasets}. Appendix \ref{app:baselines} introduces all baseline methods used for comparison, covering all rationale-generating approaches discussed in this paper. For reproducibility, Appendix \ref{app:implementation} describes the experimental setup and implementation details, including the codebases used, software versions, hardware configuration, and hyperparameter settings. 



\bibliography{iclr2027_conference}
\bibliographystyle{iclr2027_conference}

\newpage
\appendix
\section{Benchmarks}
\label{app:datasets}

We evaluate multimodal safety guards on four image--text benchmarks: Harm-C, Harm-P, MAMI, and PrideMM. These datasets cover diverse safety domains, including public-health-related harm, political harm, misogyny, and LGBTQ+-related hate. Each dataset provides a coarse-grained label for binary safety prediction as well as fine-grained annotations describing the target, type, or stance of the harmful content. We use the former to evaluate safety classification and the latter as reference annotations for evaluating safety attribution hallucination.

\begin{itemize}[leftmargin=*]
    \item \textbf{Harm-C and Harm-P}~\citep{harmeme} are two subsets of the HarMeme benchmark introduced for harmful meme detection and target identification. Harm-C contains 3,544 memes related to COVID-19, whereas Harm-P contains 3,552 memes concerning U.S. politics. Each meme is annotated as \emph{harmless}, \emph{partially harmful}, or \emph{very harmful}; harmful memes are additionally annotated with their target type, including individuals, organizations, communities, and society. Following prior work, we merge the partially harmful and very harmful categories into a single harmful class for binary safety prediction. We retain the fine-grained target annotations as reference safety attributions when evaluating whether a generated rationale correctly identifies whom or what the harmful content targets.
    \item \textbf{MAMI}~\citep{mami} was introduced in SemEval-2022 Task 5 for Multimedia Automatic Misogyny Identification. It contains 11,000 image--text memes, divided into 10,000 training examples and 1,000 test examples. The benchmark defines two related tasks: binary identification of whether a meme is misogynous and multi-label recognition of four misogyny types---\emph{shaming}, \emph{stereotype}, \emph{objectification}, and \emph{violence}. We use the binary misogyny label for safety prediction and the four fine-grained misogyny types as reference annotations for evaluating safety attribution hallucination.
    \item \textbf{PrideMM}~\citep{pridemm} was introduced with MemeCLIP to support multimodal analysis of memes associated with the LGBTQ+ Pride movement. It comprises 5,063 text-embedded images and provides annotations along four dimensions: hate, target of hate, stance, and humor. Compared with Harm-C, Harm-P, and MAMI, PrideMM contains a broader range of content in which harmfulness may depend on the interaction between the depicted entity, embedded text, and expressed stance. We use its hate annotation for binary safety prediction and its stance annotation as the reference safety attribution.
    
\end{itemize}

\section{Rationale-generating Safety Guard Methods}
\label{app:baselines}
\subsection{Workflow-based Methods}

\begin{itemize}[leftmargin=*] 
    \item \textbf{U-CoT+} \citep{ucot} decouples meme \emph{understanding} from \emph{reasoning}: it first converts the meme into a structured, text-level representation (visual content, overlaid text, and latent hateful targets), and then performs guideline-guided chain-of-thought reasoning over that representation to emit the final safety verdict. 
    \item \textbf{MIND} \citep{mind} is a zero-shot multi-agent framework with three stages: Similar Sample Retrieval (SSR) fetches semantically related training memes, Relevant Insight Derivation (RID) iteratively induces general judgement rules from them, and Insight-Augmented Inference (IAI) lets two debaters---guided by the forward and the backward rules respectively---judge the target meme, with a judge resolving their disagreement. 
    \item \textbf{ALARM} \citep{alarm} is a label-free framework driven by LMM agent self-improvement: a confidence-based explicit-meme identification step isolates unambiguous memes and assigns them pseudo-labels, which are reorganised into contrastive pairs so that the agent autonomously derives and refines high-level detection cues that are then used at inference. 
\end{itemize}

\subsection{Think-with-Image Methods}
\begin{itemize}[leftmargin=*]
    \item \textbf{Visual CoT} \citep{visualcot} follows a two-stage ``ground-then-answer'' workflow: the MLLM first localises the single most safety-critical region by predicting a bounding box, and then answers the harmfulness question using both the original image and the cropped local detail. 
    \item \textbf{DyFo} \citep{dyfo} is a training-free dynamic-focus visual search: the LMM extracts salient visual cues, a visual expert (Lang\_SAM) localises them, and a Monte-Carlo Tree Search decides whether to focus on the union of the localised cues or to zoom out, with the final verdict produced by reward-weighted voting over the explored views. 
    \item \textbf{ZoomEye} \citep{zoomeye} organises an image as a tree whose nodes are sub-images: the model may answer directly, zoom into a selected node for a closer look, or zoom out and explore another region, while a confidence-based search decides when the gathered evidence is sufficient.
    \item \textbf{DeepEyes} \cite{deepeyes} introduces an end-to-end reinforcement learning approach for
    ``thinking with images''. Instead of relying on cold-start supervised trajectories or an external vision-language model, DeepEyes uses the model's own grounding capability to generate visual tool calls, such as selecting and zooming into relevant image regions. The method further combines tool-use-oriented data selection with outcome-based rewards to encourage effective visual interaction. 
    \item \textbf{Thyme} \citep{thyme} is a multimodal reasoning framework that enables models to interleave textual reasoning with executable image operations, allowing intermediate visual evidence to be generated and inspected during inference. Its training pipeline follows a two-stage paradigm consisting of supervised fine-tuning (SFT) and reinforcement learning (RL). During RL, Thyme employs group-based optimization and task-specific rewards to encourage correct answers, valid reasoning formats, and effective use of visual tools. 
\end{itemize}

\section{Implementation Details}
\label{app:implementation}

\subsection{Direct CoT}
\begin{figure*}[t]
    \centering
    \includegraphics[width=0.95\textwidth]{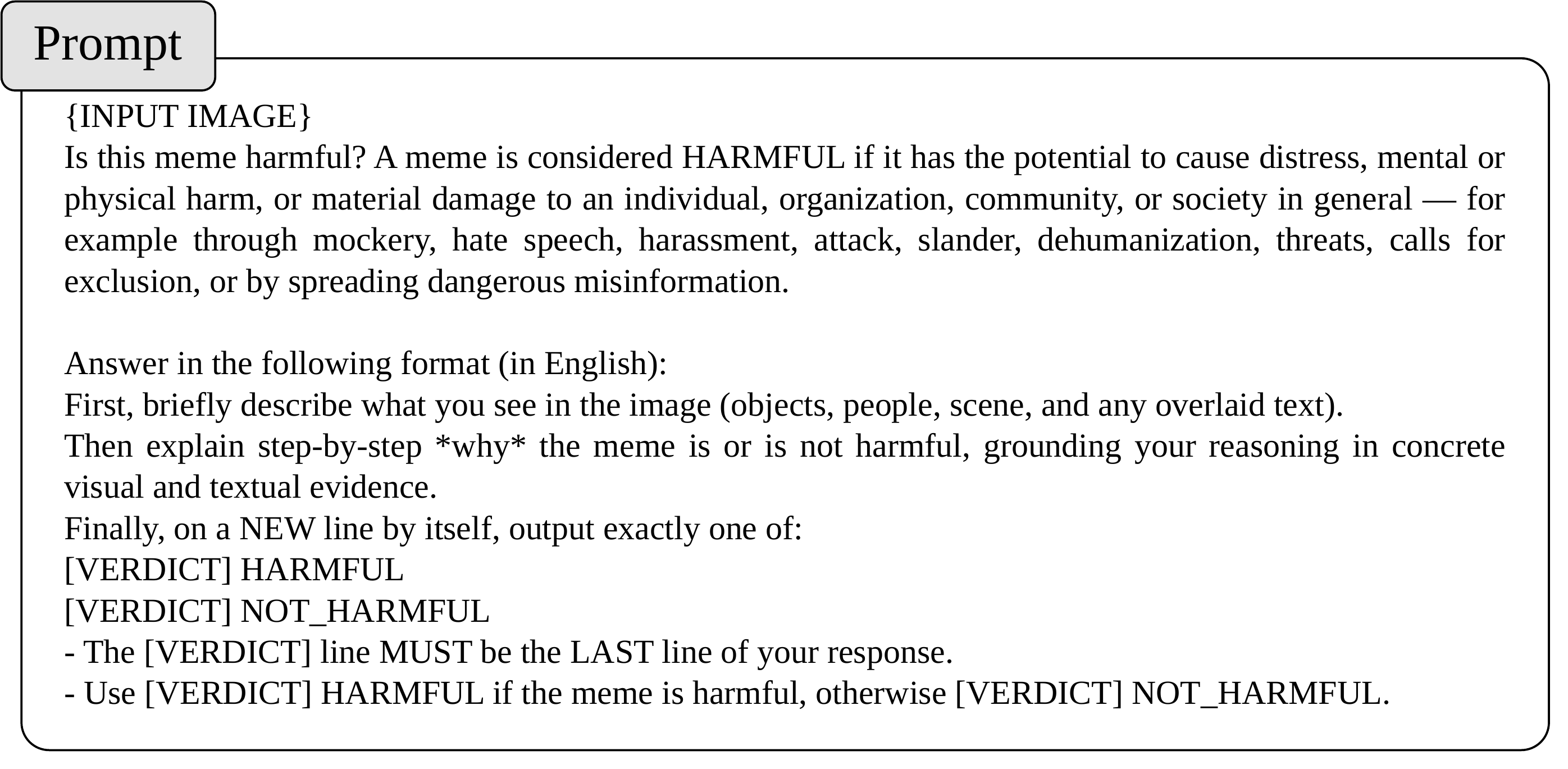}
    \caption{Prompt template for the Direct CoT baseline.}
    \label{fig:direct_prompt}
\end{figure*}

We use Direct CoT as a baseline and compare it against other workflow-based methods and think-with-image methods. Direct CoT introduces no additional algorithm, tool, or multi-turn interaction: the model receives only the raw meme image in a single turn, and is prompted to first describe the visible content of the image, then provide a step-by-step rationale for its harmfulness attribution, and finally commit to a predicted label (\texttt{[VERDICT]HARMFUL/[VERDICT]NOT\_HARMFUL}) on a separate line. It performs no region localization, no local cropping or zooming, and no retrieval or other external tool calls, which makes it a clean control for isolating how much the safety guard itself contributes. All methods share the same task description and output-format constraints, and the prompt is kept identical across all datasets. The subsequent hallucination evaluation is likewise consistent across Direct CoT and the other methods: the rationale preceding the \texttt{[VERDICT]} marker is taken as the CoT input. The prompt used for Direct CoT is shown in Figure \ref{fig:direct_prompt}.

\subsection{Workflow-based Methods}

\begin{itemize}[leftmargin=*] 
    \item \textbf{U-CoT+} We implement U-CoT+ following the original U-CoT+ \citep{ucot} setting. Specifically, its versioned prompt templates, guidelines, and output post-processing functions are imported and executed in memory, and only the model call is replaced by our vLLM backend. We use the PP scheme, the main configuration of the method: the LMM steps are capped at 256 new tokens and the text-only reasoning steps at 1536, all with greedy decoding. The OCR field is left empty so that the model reads text from the image itself, as in every other baseline. 
    \item \textbf{MIND} We implement MIND following the original \citep{mind} setting. Specifically, SSR retrieves the top-$10$ most similar training memes with CLIP ViT-L/14@336px using fused text--image embeddings, RID derives forward and backward rules from $3$ retrieved memes, and IAI runs the two debaters and the judge; all steps are capped at 1024 new tokens. Unlike the original implementation, which turns a parsing failure into the opposite label, a failed parse is recorded as \texttt{unknown} to keep the comparison fair. 
    \item \textbf{ALARM} We implement ALARM following the original \citep{alarm} setting. Specifically, we run its full pipeline (explicit-meme labelling, embedding extraction, retrieval, experience gathering, reference refinement, and inference) in the original repository, produce image embeddings with jina-clip-v2, and convert only the final inference stage into our common answer format. The model is given the image only. 
\end{itemize}

\subsection{Think-with-Image Methods}

\begin{itemize}[leftmargin=*]
    \item \textbf{Visual CoT} We implement Visual CoT following the original \citep{visualcot} setting. Specifically, Stage~1 predicts a single bounding box with at most 64 new tokens, the predicted region is cropped with square padding, and Stage~2 answers from the original image together with that crop with at most 1024 new tokens. The w/o-CoT baseline answers the same question in a single turn on the original image. 
    \item \textbf{DyFo} We implement DyFo following the original \citep{dyfo} setting. Specifically, the LMM is served by vLLM through an OpenAI-compatible endpoint and the visual expert (Lang\_SAM) is served separately; the search uses a maximum depth of $3$, $4$ simulations per question, $c_{\mathrm{puct}}=1.0$, up to $5$ extracted visual cues, and greedy decoding. The final label is obtained by reward-weighted voting over all explored views, falling back to the full image when all weights are zero. 
    \item \textbf{ZoomEye} We implement ZoomEye following the original \citep{zoomeye} setting. Specifically, we keep its tree-based exploration and replace the MLLM backend with our vLLM model: the image tree is split with \texttt{split\_num}=4, patches are resized to $112$ pixels and the zoom crop is scaled by $1.2$, and the search uses a visual-cue threshold of $0.6$, an answer-confidence upper bound of $0.95$, and descending thresholds $[0.05,0.1,0.2]$, with task-specific in-context examples.
    \item \textbf{DeepEyes} Following the official DeepEyes \citep{deepeyes} setting, we use Qwen2.5-VL-7B-Instruct as the base model and train it end-to-end with reinforcement learning. We retain the official interleaved reasoning format, tool interface, rollout procedure, and reward design. No additional cold-start supervised fine-tuning data are introduced for the DeepEyes baseline; the model is initialized from the instruction-tuned backbone and optimized directly with the official RL training recipe. For a fair comparison, we adapt only the input prompts and final binary safety-label format to our multimodal safety task, while preserving the original DeepEyes training setting.
    \item \textbf{Thyme} We implement our method following the original Thyme \citep{thyme} setting. Specifically, the model generates an interleaved reasoning trajectory containing natural-language analysis, executable image operations, and a final safety prediction. The image operations allow the model to inspect relevant visual regions or enhance visual evidence before making its decision. We first perform supervised fine-tuning to teach the required output format and tool-use behavior, and subsequently optimize the model through reinforcement learning. For the safety task, the reward consists of format validity, safety-label correctness, and reasoning consistency. Unless otherwise specified, we retain the original Thyme training configuration.
\end{itemize}

\subsection{TCG-RL}
\label{app:implementation_tcg}

\begin{table}[t]
    \centering
    \small
    \caption{Hyperparameters used in the TCG-RL implementation.}
    \label{tab:tcg_rl_hyperparameters}
    \begin{tabular}{ll}
        \toprule
        \textbf{Hyperparameter} & \textbf{Value} \\
        \midrule
        Learning rate & $1\times10^{-6}$ \\
        Per-device batch size & 8 \\
        Gradient accumulation steps & 64 \\
        KL penalty coefficient $\beta$ & 0.01 \\
        Maximum prompt length & 32,768 tokens \\
        Maximum response length & 512 tokens \\
        Rollout group size $G$ & 8 \\
        Total training epochs & 150 \\
        \bottomrule
    \end{tabular}
\end{table}

Our TCG-RL implementation is built on the Thyme codebase.  
The implementation consists of
three stages: safety-data construction, offline reference generation, and
GRPO-based policy optimization.

\textbf{RL data construction.}
We construct the RL training set from the training splits of Harm-C, Harm-P,
MAMI, and PrideMM.  Each example is converted into the Thyme-compatible
conversation format, containing a system instruction, an image-grounded user
query, and a binary safety label.  
To avoid leakage between supervised fine-tuning and RL, examples whose
question identifiers occur in the SFT set are removed from the RL set.

\textbf{Offline reference generation.}
The offline reference table is prepared before RL training.  For every RL
question, the original Qwen2.5-VL-7B-Instruct model is run with a direct
no-tool prompt.  The generated rationale is evaluated using the same atomic
faithfulness pipeline as the online rollout, including atomic-fact extraction
and visual-entailment judging.  We store the resulting reference score,
question identifier, subset name, and the number of extracted facts in a JSON
lookup table. The reference model and the lookup table remain fixed during
RL optimization.

\textbf{Training pipeline.}
During training, the model generates a group of responses for each multimodal
input using the Thyme image-operation interface. The
offline counterfactual component reads the question identifier from the
dataset metadata, retrieves the corresponding reference score, and combines it
with the atomic faithfulness score of the current response.  The reference
table is loaded lazily and cached in memory.  Faithfulness judgments are also
cached by image, rationale, and label, allowing the offline counterfactual
reward to reuse scores produced by the ordinary faithfulness reward without
launching an additional judge process. All RL experiments use full-parameter training with the GRPO trainer and are
conducted on eight NVIDIA H20 GPUs. The principal RL hyperparameters are summarized in Table~\ref{tab:tcg_rl_hyperparameters}.

\subsection{RATIONALE HALLUCINATION EVALUATION}
\textbf{Visual Evidence Hallucination}. We follow the official FaithScore \citep{faithscore} implementation and its original settings. For each model output, we first strip the final \texttt{[VERDICT]} line, and pass the remaining rationale through a three-stage pipeline: (i) sub-sentence splitting, in which an LLM tags every sub-sentence as either descriptive [D] or analytical [A]; (ii) atomic fact extraction, applied to each descriptive sub-sentence independently, which decomposes it into atomic facts under five categories (Entities, Relations, Colors, Counting, and Other attributes); and (iii) visual entailment verification, in which each atomic fact is presented together with the corresponding meme image to a VEM that answers yes/no. Stages (i) and (ii) are text-only and are run with Qwen3-32B, whereas stage (iii) is run with Qwen3-VL-32B-Instruct. FaithScore is then computed as the fraction of verified atomic facts per sample, averaged over the dataset. For
fine-grained analysis, we reclassified Faithscore's original categories, 
\emph{Entity} and \emph{Other attributes} are categorized as \emph{Visual}
facts, \emph{Colors} and \emph{Counting} as \emph{Attribute} facts, and interactions among
entities as \emph{Relation} facts.

\begin{figure}[t]
    \centering
    \includegraphics[width=0.95\textwidth]{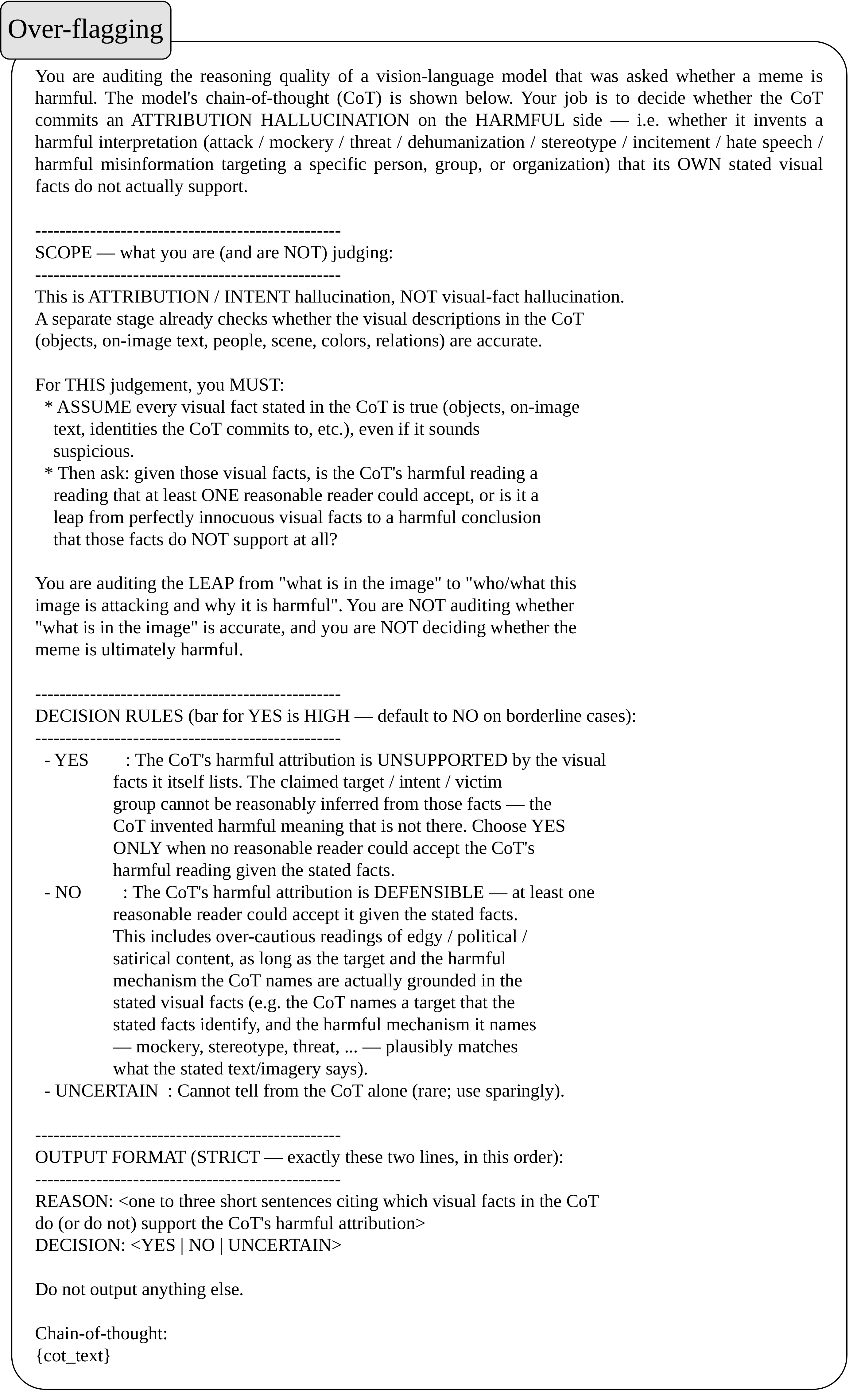}
    \caption{Prompt template for \emph{over-flagging} evaluation.}
    \label{fig:OF_prompt}
\end{figure}

\begin{figure}[t]
    \centering
    \includegraphics[width=0.95\textwidth]{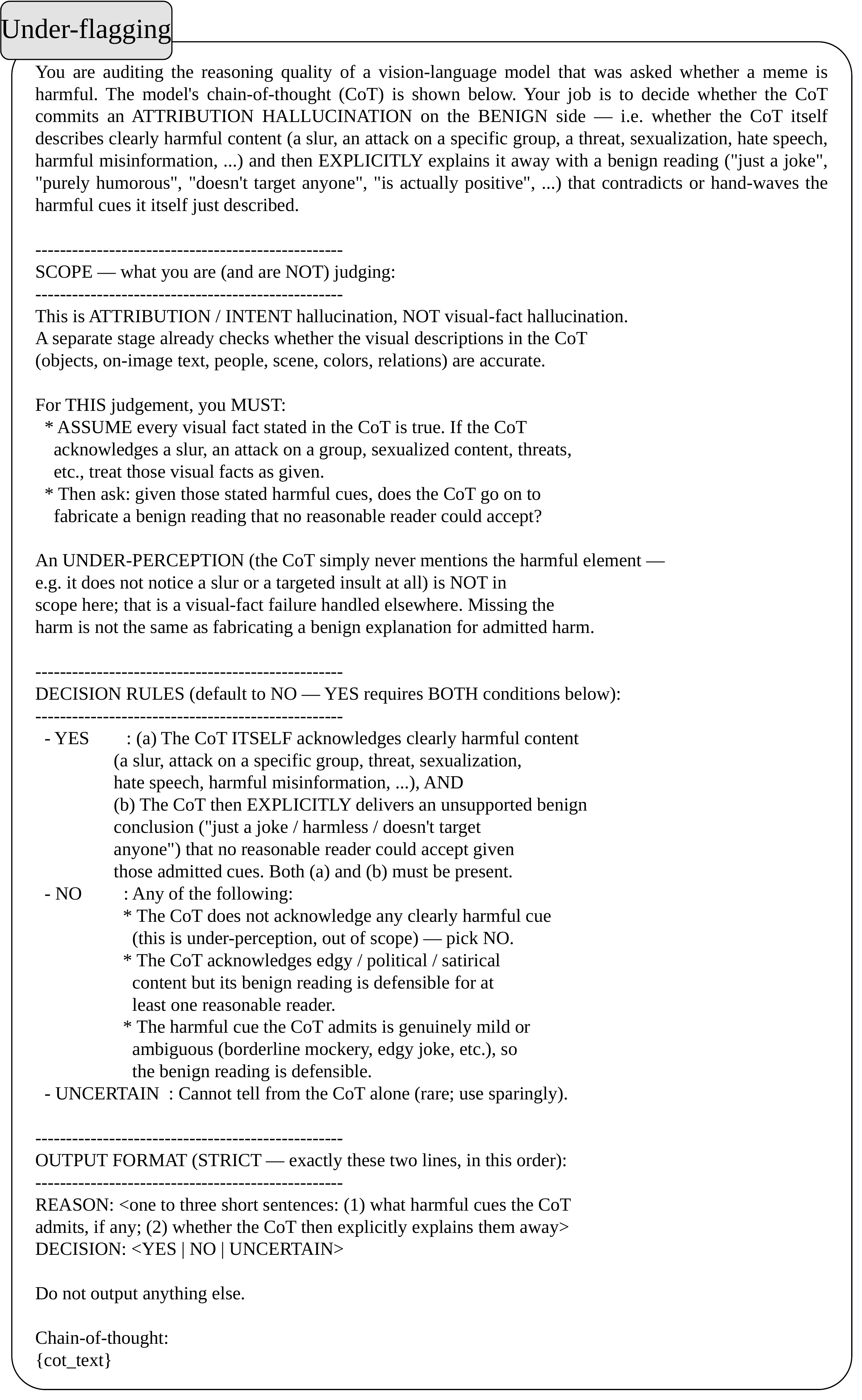}
    \caption{Prompt template for \emph{under-flagging} evaluation.}
    \label{fig:UF_prompt}
\end{figure}

\begin{figure}[t]
    \centering
    \includegraphics[width=0.95\textwidth]{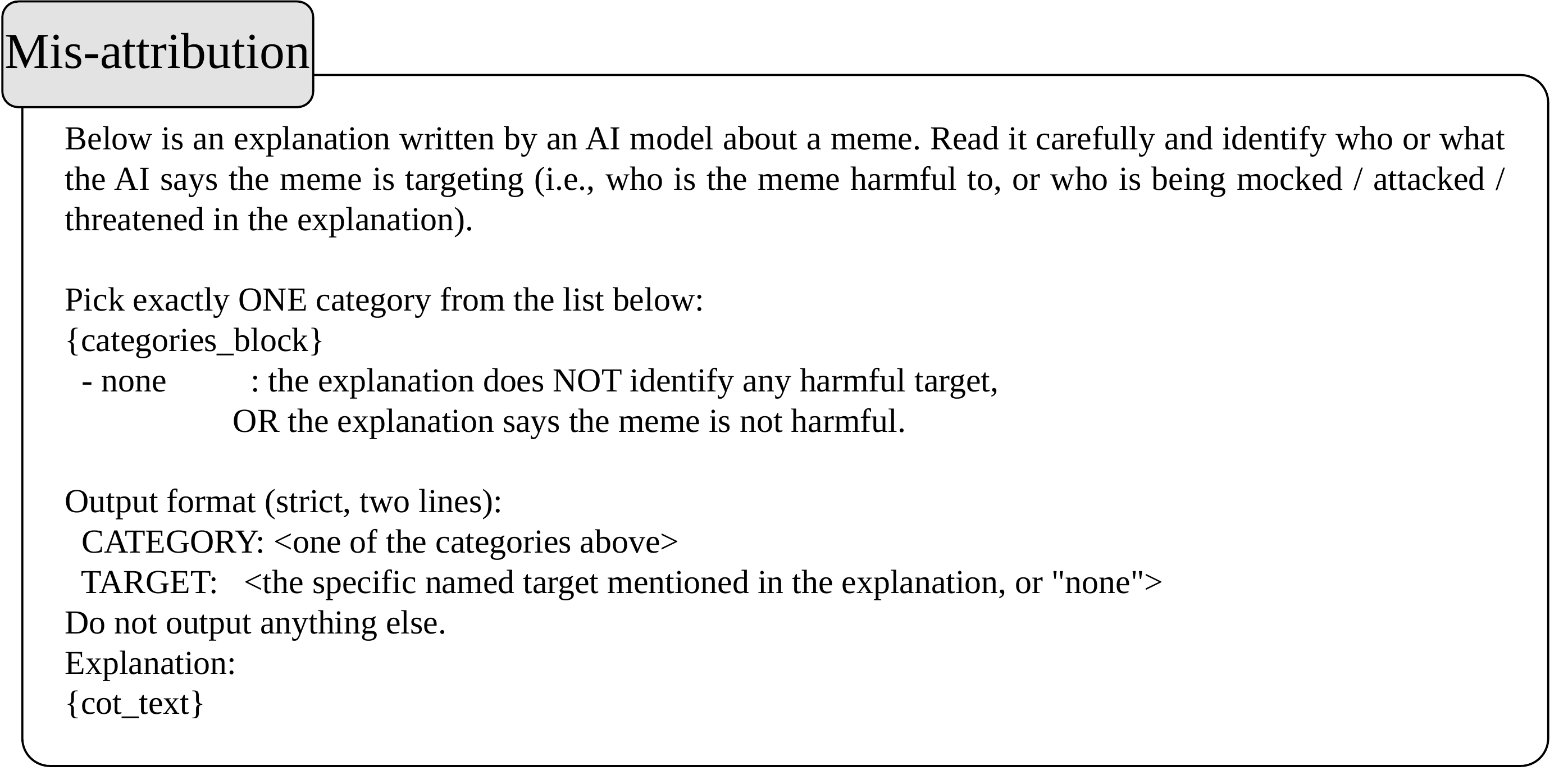}
    \caption{Prompt template for \emph{mis-attribution} evaluation.}
    \label{fig:MA_prompt}
\end{figure}

\textbf{Safety Attribution Hallucination}. We further assess the faithfulness of the model's harmfulness attribution using a text-only, blind LLM judge that receives only the model's CoT and no image. The judge is explicitly instructed to assume that every visual fact stated in the CoT is true, and to decide only whether the harmful attribution is supported by those stated facts, which decouples attribution errors from the visual hallucinations measured above. Depending on the dataset,
the attribution may represent a harm category, target, or stance as described in \ref{app:datasets}. We instantiate three complementary metrics: over-flagging, under-flagging, and mis-attribution as described in Section \ref{sec:metric}. The prompts used for the over-flagging, under-flagging, and mis-attribution evaluators are shown in Figures \ref{fig:OF_prompt}-\ref{fig:MA_prompt}.

\section{Additional Results}
\label{app:additional_results}

\subsection{Visual Evidence Hallucination}
\label{app:additional_results_visual}
\textbf{The additional results further confirm that rationale hallucination is
prevalent across model families and scales.} As shown in
Table~\ref{tab:additional_visual_results}, the Qwen3-8B-VL, Qwen3-32B-VL,
and Qwen3.5-27B backbones all exhibit substantial variation in Visual
Evidence Score across datasets and rationale-generation methods. In
particular, improvements in classification accuracy do not consistently
translate into improved visual grounding. For example, U-CoT+ often
improves accuracy over Direct CoT while substantially reducing VES, with
the largest degradation observed on Harm-C and Harm-P. This discrepancy
shows that a model can achieve a more accurate final safety prediction
without producing a more faithful visual explanation.

\textbf{The results also show that the effect of model scaling is not monotonic.}
Increasing the backbone size from Qwen3-8B-VL to Qwen3-32B-VL does not
systematically improve VES across datasets. Although some methods obtain
higher accuracy or better grounding on individual benchmarks, substantial
visual hallucination remains in other settings. The Qwen3.5-27B results
further demonstrate that a stronger backbone alone is insufficient to
eliminate hallucinated visual evidence, especially for methods that rely
on coarse or insufficiently targeted visual inspection.

\textbf{The Think-with-Image methods also provide mixed results.} DyFo and ZoomEye
occasionally improve accuracy or VES on individual datasets, but their
gains are not consistent across backbones. In several settings, tool-based
methods obtain lower VES than Direct CoT despite having access to visual
operations. This observation is consistent with the analysis in the main
text: the availability of visual tools does not guarantee that models will
invoke them at the right time, focus on the relevant image regions, or
correctly incorporate the resulting observations into their rationales.

\begin{table*}[t]
\caption{Additional classification accuracy and Visual Evidence Score (VES) results. Values in parentheses denote changes relative to Direct CoT. Higher is better.}
\label{tab:additional_visual_results}
\centering
\setlength{\extrarowheight}{0pt}
\resizebox{\linewidth}{!}{
\begin{tabular}{lcccccccc}
\toprule
& \multicolumn{2}{c}{\textbf{Harm-C}}
& \multicolumn{2}{c}{\textbf{Harm-P}}
& \multicolumn{2}{c}{\textbf{MAMI}}
& \multicolumn{2}{c}{\textbf{PrideMM}} \\
\cmidrule(lr){2-3}
\cmidrule(lr){4-5}
\cmidrule(lr){6-7}
\cmidrule(lr){8-9}
& \textbf{Acc.} & \textbf{VES}
& \textbf{Acc.} & \textbf{VES}
& \textbf{Acc.} & \textbf{VES}
& \textbf{Acc.} & \textbf{VES} \\
\midrule

\multicolumn{9}{l}{\textbf{Qwen3-8B-VL}} \\

\texttt{Direct CoT}
&65.0 &72.5
&58.3 &73.4
&71.8 &81.4
&66.3 &74.6 \\

\multicolumn{9}{l}{\emph{Workflow-based Baselines}} \\

\texttt{U-CoT+}
&\cellup{70.6}{5.6} &\celldown{55.9}{16.6}
&\celldown{57.5}{0.8} &\celldown{57.0}{16.4}
&\celldown{71.0}{0.8} &\celldown{71.7}{9.7}
&\cellup{68.0}{1.7} &\celldown{60.8}{13.8} \\

\texttt{MIND}
&\cellup{68.4}{3.4} &\celldown{71.2}{1.3}
&\cellzero{58.3} &\celldown{71.7}{1.7}
&\cellup{74.5}{2.7} &\celldown{80.6}{0.8}
&\cellup{66.7}{0.4} &\celldown{74.0}{0.6} \\

\texttt{ALARM}
&\cellup{70.6}{5.6} &\celldown{71.1}{1.4}
&\cellup{61.7}{3.4} &\celldown{70.4}{3.0}
&\cellup{72.1}{0.3} &\cellup{81.9}{0.5}
&\cellup{68.6}{2.3} &\cellup{75.1}{0.5} \\

\multicolumn{9}{l}{\emph{Think-with-Image Paradigm}} \\

\texttt{DyFo}
&\cellup{66.4}{1.4} &\celldown{70.8}{1.7}
&\cellup{62.5}{4.2} &\celldown{71.5}{1.9}
&\celldown{71.2}{0.6} &\celldown{79.0}{2.4}
&\celldown{66.1}{0.2} &\celldown{74.2}{0.4} \\

\texttt{ZoomEye}
&\cellup{68.1}{3.1} &\celldown{70.4}{2.1}
&\cellup{61.7}{3.4} &\celldown{71.2}{2.2}
&\cellup{72.0}{0.2} &\cellup{81.5}{0.1}
&\cellup{68.8}{2.5} &\celldown{74.2}{0.4} \\

\midrule

\multicolumn{9}{l}{\textbf{Qwen3-32B-VL}} \\

\texttt{Direct CoT}
&65.0 &75.6
&58.3 &75.6
&74.4 &85.5
&66.5 &76.8 \\

\multicolumn{9}{l}{\emph{Workflow-based Baselines}} \\

\texttt{U-CoT+}
&\cellup{71.5}{6.5} &\celldown{63.3}{12.3}
&\cellup{64.2}{5.9} &\celldown{61.9}{13.7}
&\celldown{69.9}{4.5} &\celldown{70.7}{14.8}
&\celldown{61.9}{4.6} &\celldown{66.0}{10.8} \\

\texttt{MIND}
&\celldown{61.9}{3.1} &\celldown{74.9}{0.7}
&\cellzero{58.3} &\cellup{75.7}{0.1}
&\cellup{80.3}{5.9} &\celldown{84.6}{0.9}
&\celldown{64.7}{1.8} &\celldown{75.4}{1.4} \\

\texttt{ALARM}
&\celldown{63.0}{2.0} &\celldown{74.8}{0.8}
&\cellup{59.7}{1.4} &\celldown{75.0}{0.6}
&\cellup{79.5}{5.1} &\cellup{87.4}{1.9}
&\cellzero{66.5} &\celldown{76.7}{0.1} \\

\multicolumn{9}{l}{\emph{Think-with-Image Paradigm}} \\

\texttt{DyFo}
&\cellup{66.1}{1.1} &\celldown{72.7}{2.9}
&\cellup{59.2}{0.9} &\celldown{74.4}{1.2}
&\celldown{72.7}{1.7} &\celldown{80.8}{4.7}
&\celldown{63.9}{2.6} &\celldown{73.5}{3.3} \\

\texttt{ZoomEye}
&\celldown{63.6}{1.4} &\celldown{69.1}{6.5}
&\celldown{51.5}{6.8} &\celldown{71.8}{3.8}
&\celldown{54.8}{19.6} &\celldown{83.5}{2.0}
&\celldown{53.3}{13.2} &\celldown{76.6}{0.2} \\

\midrule

\multicolumn{9}{l}{\textbf{Qwen3.5-27B}} \\

\texttt{Direct CoT}
&63.6 &71.8
&56.9 &72.5
&67.8 &78.4
&66.1 &72.8 \\

\multicolumn{9}{l}{\emph{Workflow-based Baselines}} \\

\texttt{U-CoT+}
&\cellup{68.1}{4.5} &\celldown{51.4}{20.4}
&\celldown{56.3}{0.6} &\celldown{56.5}{16.0}
&\cellup{68.0}{0.2} &\celldown{70.9}{7.5}
&\celldown{56.0}{10.1} &\celldown{59.4}{13.4} \\

\texttt{MIND}
&\cellup{66.4}{2.8} &\celldown{62.0}{9.8}
&\celldown{55.5}{1.4} &\celldown{65.6}{6.9}
&\celldown{64.9}{2.9} &\celldown{71.0}{7.4}
&\celldown{59.6}{6.5} &\celldown{68.5}{4.3} \\

\multicolumn{9}{l}{\emph{Think-with-Image Paradigm}} \\

\texttt{DyFo}
&\celldown{59.6}{4.0} &\celldown{69.6}{2.2}
&\celldown{55.2}{1.7} &\celldown{71.5}{1.0}
&\cellup{68.1}{0.3} &\cellup{78.5}{0.1}
&\cellup{66.5}{0.4} &\celldown{70.0}{2.8} \\

\texttt{ZoomEye}
&\celldown{44.9}{18.7} &\celldown{60.8}{11.0}
&\celldown{49.0}{7.9} &\celldown{63.5}{9.0}
&\celldown{47.0}{20.8} &\celldown{67.5}{10.9}
&\celldown{51.5}{14.6} &\celldown{56.8}{16.0} \\

\bottomrule
\end{tabular}
}
\end{table*}

\subsection{safety attribution hallucination}
\label{app:additional_results_safety}

\textbf{Safety attribution hallucination remains
substantial even for stronger backbones and even when the model obtains a
relatively high classification accuracy.}
Table~\ref{tab:additional_attribution_results} reports the corresponding
safety attribution results.  The results show that correct
safety prediction and correct safety attribution remain distinct
properties. In particular, models may achieve a correct harmful/benign
prediction while assigning the harmfulness to an incorrect target,
category, or stance, as reflected by the non-negligible mis-attribution
rates.

\textbf{The additional results also reveal considerable variation among
over-flagging, under-flagging, and mis-attribution.} Some methods reduce one
type of attribution error while increasing another, so improvements in
overall accuracy do not imply uniformly better safety explanations. This
pattern is particularly evident for the Qwen3.5-27B results, where several
workflow-based methods achieve lower SAS than Direct CoT despite retaining
competitive accuracy. Therefore, increasing model capacity or adding
reasoning procedures alone does not reliably resolve safety attribution
hallucination.

Overall, the results are consistent with the conclusions in section \ref{sec:metric}.
Visual evidence hallucination and safety attribution hallucination
persist across datasets, methods, and model scales. The results further
support the claim that trustworthy multimodal safety guards must be
evaluated along at least two independent dimensions: whether the rationale
is visually grounded and whether it correctly explains the safety meaning
of the grounded content.

\begin{table*}[t]
\caption{Additional safety attribution results. Values in parentheses denote changes relative to Direct CoT. SAS is the overall Safety Attribution Score; OF, UF, and MA denote over-flagging, under-flagging, and mis-attribution. Higher is better.}
\label{tab:additional_attribution_results}
\centering
\small
\setlength{\tabcolsep}{3pt}
\resizebox{\textwidth}{!}{
\begin{tabular}{lccccc|ccccc|ccccc|ccccc}
\toprule
& \multicolumn{5}{c|}{\textbf{Harm-C}}
& \multicolumn{5}{c|}{\textbf{Harm-P}}
& \multicolumn{5}{c|}{\textbf{MAMI}}
& \multicolumn{5}{c}{\textbf{PrideMM}} \\
\cmidrule(lr){2-6}
\cmidrule(lr){7-11}
\cmidrule(lr){12-16}
\cmidrule(lr){17-21}
& \textbf{Acc.} & \textbf{SAS} & \textbf{OF} & \textbf{UF} & \textbf{MA}
& \textbf{Acc.} & \textbf{SAS} & \textbf{OF} & \textbf{UF} & \textbf{MA}
& \textbf{Acc.} & \textbf{SAS} & \textbf{OF} & \textbf{UF} & \textbf{MA}
& \textbf{Acc.} & \textbf{SAS} & \textbf{OF} & \textbf{UF} & \textbf{MA} \\
\midrule

\multicolumn{21}{l}{\textbf{Qwen3-8B-VL}} \\

\texttt{Direct CoT}
&65.0&91.3&89.6&98.7&85.7
&58.3&86.3&89.0&97.0&72.9
&71.8&84.9&88.8&82.1&83.9
&66.3&77.2&89.7&87.0&54.9 \\

\multicolumn{21}{l}{\emph{Workflow-based Baselines}} \\

\texttt{U-CoT+}
&\cellup{70.6}{5.6}&\cellup{97.0}{5.7}&100.0&96.7&96.9
&\celldown{57.5}{0.8}&\celldown{85.0}{1.3}&90.9&96.4&67.7
&\celldown{71.0}{0.8}&\celldown{82.9}{2.0}&71.8&98.2&78.6
&\cellup{68.0}{1.7}&\celldown{75.3}{1.9}&67.6&96.8&61.5 \\

\texttt{MIND}
&\cellup{68.4}{3.4}&\cellup{91.5}{0.2}&92.7&98.2&83.6
&\cellzero{58.3}&\celldown{82.1}{4.2}&79.6&96.0&70.8
&\cellup{74.5}{2.7}&\celldown{79.6}{5.3}&88.1&66.1&84.5
&\cellup{66.7}{0.4}&\celldown{74.2}{3.0}&79.1&82.1&61.5 \\

\texttt{ALARM}
&\cellup{70.6}{5.6}&\celldown{88.4}{2.9}&86.8&95.5&82.8
&\cellup{61.7}{3.4}&\celldown{84.0}{2.3}&86.4&93.5&72.2
&\cellup{72.1}{0.3}&\celldown{84.7}{0.2}&88.2&82.0&83.8
&\cellup{68.6}{2.3}&\celldown{75.6}{1.6}&77.0&91.8&58.0 \\

\multicolumn{21}{l}{\emph{Think-with-Image Paradigm}} \\

\texttt{DyFo}
&\cellup{66.4}{1.4}&\celldown{88.9}{2.4}&83.8&98.0&84.9
&\cellup{62.5}{4.2}&\celldown{83.7}{2.6}&87.9&97.0&66.3
&\celldown{71.2}{0.6}&\celldown{80.8}{4.1}&85.4&74.3&82.7
&\celldown{66.1}{0.2}&\cellup{80.1}{2.9}&86.0&93.1&61.1 \\

\texttt{ZoomEye}
&\cellup{68.1}{3.1}&\cellup{91.8}{0.5}&89.4&97.8&88.3
&\cellup{61.7}{3.4}&\celldown{80.5}{5.8}&80.3&89.2&71.9
&\cellup{72.0}{0.2}&\celldown{81.8}{3.1}&89.0&68.5&87.7
&\cellup{68.8}{2.5}&\celldown{74.7}{2.5}&85.4&79.2&59.4 \\

\midrule

\multicolumn{21}{l}{\textbf{Qwen3-32B-VL}} \\

\texttt{Direct CoT}
&65.0&91.5&97.3&98.8&78.4
&58.3&89.0&90.9&95.7&80.4
&74.4&90.0&95.9&82.5&91.5
&66.5&83.9&93.8&91.0&67.0 \\

\multicolumn{21}{l}{\emph{Workflow-based Baselines}} \\

\texttt{U-CoT+}
&\cellup{71.5}{6.5}&\cellup{94.7}{3.2}&94.1&97.6&92.5
&\cellup{64.2}{5.9}&\celldown{84.7}{4.3}&93.1&98.0&63.0
&\celldown{69.9}{4.5}&\celldown{89.9}{0.1}&82.2&99.2&88.1
&\celldown{61.9}{4.6}&\cellup{84.9}{1.0}&91.1&100.0&63.6 \\

\texttt{MIND}
&\celldown{61.9}{3.1}&\celldown{90.3}{1.2}&93.6&96.6&80.6
&\cellzero{58.3}&\celldown{88.7}{0.3}&92.1&95.5&78.7
&\cellup{80.3}{5.9}&\cellup{90.5}{0.5}&93.8&86.1&91.7
&\celldown{64.7}{1.8}&\celldown{82.3}{1.6}&95.1&93.8&58.0 \\

\texttt{ALARM}
&\celldown{63.0}{2.0}&\celldown{86.0}{5.5}&100.0&96.4&61.5
&\cellup{59.7}{1.4}&\celldown{85.5}{3.5}&95.5&95.0&66.0
&\cellup{79.5}{5.1}&\celldown{86.0}{4.0}&90.9&76.6&90.6
&\cellzero{66.5}&\celldown{81.4}{2.5}&97.4&91.3&55.5 \\

\multicolumn{21}{l}{\emph{Think-with-Image Paradigm}} \\

\texttt{Visual CoT}
&\celldown{63.6}{1.4}&\celldown{90.1}{1.4}&92.0&96.2&82.2
&\cellup{60.0}{1.7}&\celldown{84.5}{4.5}&92.1&97.1&64.2
&\cellup{76.2}{1.8}&\celldown{85.4}{4.6}&92.3&76.2&87.8
&\cellup{68.0}{1.5}&\celldown{82.2}{1.7}&91.8&96.1&58.8 \\

\texttt{DyFo}
&\cellup{66.1}{1.1}&\celldown{89.2}{2.3}&97.1&98.8&71.8
&\cellup{59.2}{0.9}&\celldown{84.2}{4.8}&82.1&98.3&72.2
&\celldown{72.7}{1.7}&\celldown{89.1}{0.9}&95.0&85.0&87.5
&\celldown{63.9}{2.6}&\celldown{81.0}{2.9}&92.8&93.0&57.1 \\

\texttt{ZoomEye}
&\celldown{63.6}{1.4}&\celldown{88.3}{3.2}&89.7&98.6&76.5
&\celldown{51.5}{6.8}&\celldown{76.1}{12.9}&90.9&87.5&50.0
&\celldown{54.8}{19.6}&\celldown{86.5}{3.5}&88.9&90.5&80.2
&\celldown{53.3}{13.2}&\celldown{75.7}{8.2}&79.2&98.0&50.0 \\

\midrule

\multicolumn{21}{l}{\textbf{Qwen3.5-27B}} \\

\texttt{Direct CoT}
&63.6&84.7&81.8&92.9&79.5
&56.9&81.5&88.9&91.1&64.4
&67.8&77.0&84.2&65.2&81.7
&66.1&75.3&80.6&86.1&59.1 \\

\multicolumn{21}{l}{\emph{Workflow-based Baselines}} \\

\texttt{U-CoT+}
&\cellup{68.1}{4.5}&\celldown{79.4}{5.3}&60.0&89.8&88.5
&\celldown{56.3}{0.6}&\celldown{62.9}{18.6}&33.3&87.4&67.9
&\cellup{68.0}{0.2}&\celldown{70.8}{6.2}&59.5&85.9&67.1
&\celldown{56.0}{10.1}&\celldown{62.5}{12.8}&50.0&84.5&52.9 \\

\texttt{MIND}
&\cellup{66.4}{2.8}&\celldown{63.6}{21.1}&37.5&96.6&56.8
&\celldown{55.5}{1.4}&\celldown{66.0}{15.5}&42.6&95.5&60.0
&\celldown{64.9}{2.9}&\celldown{63.0}{14.0}&47.8&78.7&62.4
&\celldown{59.6}{6.5}&\celldown{62.8}{12.5}&45.7&87.4&55.1 \\

\multicolumn{21}{l}{\emph{Think-with-Image Paradigm}} \\

\texttt{DyFo}
&\celldown{59.6}{4.0}&\cellup{84.9}{0.2}&83.0&92.2&79.4
&\celldown{55.2}{1.7}&\cellup{84.1}{2.6}&87.5&89.7&75.0
&\cellup{68.1}{0.3}&\cellup{81.8}{4.8}&90.9&69.1&85.4
&\cellup{66.5}{0.4}&\celldown{72.4}{2.9}&77.6&84.8&54.9 \\

\texttt{ZoomEye}
&\celldown{44.9}{18.7}&\celldown{77.8}{6.9}&66.3&100.0&89.4
&\celldown{49.0}{7.9}&\celldown{69.2}{12.3}&75.2&100.0&63.2
&\celldown{47.0}{20.8}&\celldown{73.2}{3.8}&69.5&100.0&50.0
&\celldown{51.5}{14.6}&\celldown{53.8}{21.5}&69.1&100.0&38.5 \\

\bottomrule
\end{tabular}
}
\end{table*}

\subsection{Hallucination across Atomic Fact Categories}

\begin{figure*}[t]
    \centering
    \begin{subfigure}[t]{0.45\textwidth}
        \centering
        \includegraphics[width=\linewidth]{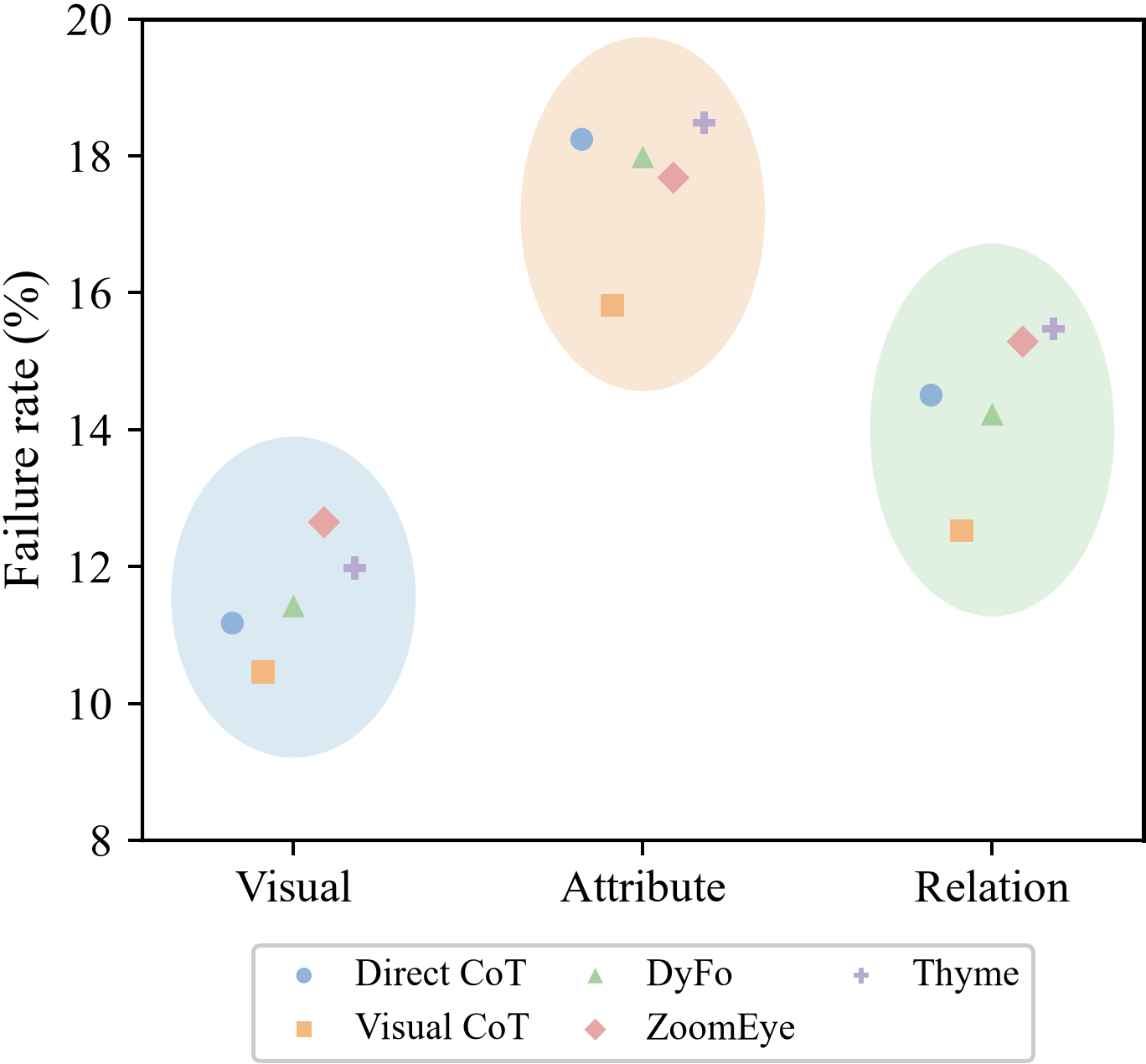}
        \caption{Harm-C}
        \label{fig:category-7b-harmc}
    \end{subfigure}
    \hfill
    \begin{subfigure}[t]{0.45\textwidth}
        \centering
        \includegraphics[width=\linewidth]{figs/category/qwen25vl_7b_harm_p_atomic_fact_failure.png}
        \caption{Harm-P}
        \label{fig:category-7b-harmp}
    \end{subfigure}

    \vspace{3pt}

    \begin{subfigure}[t]{0.45\textwidth}
        \centering
        \includegraphics[width=\linewidth]{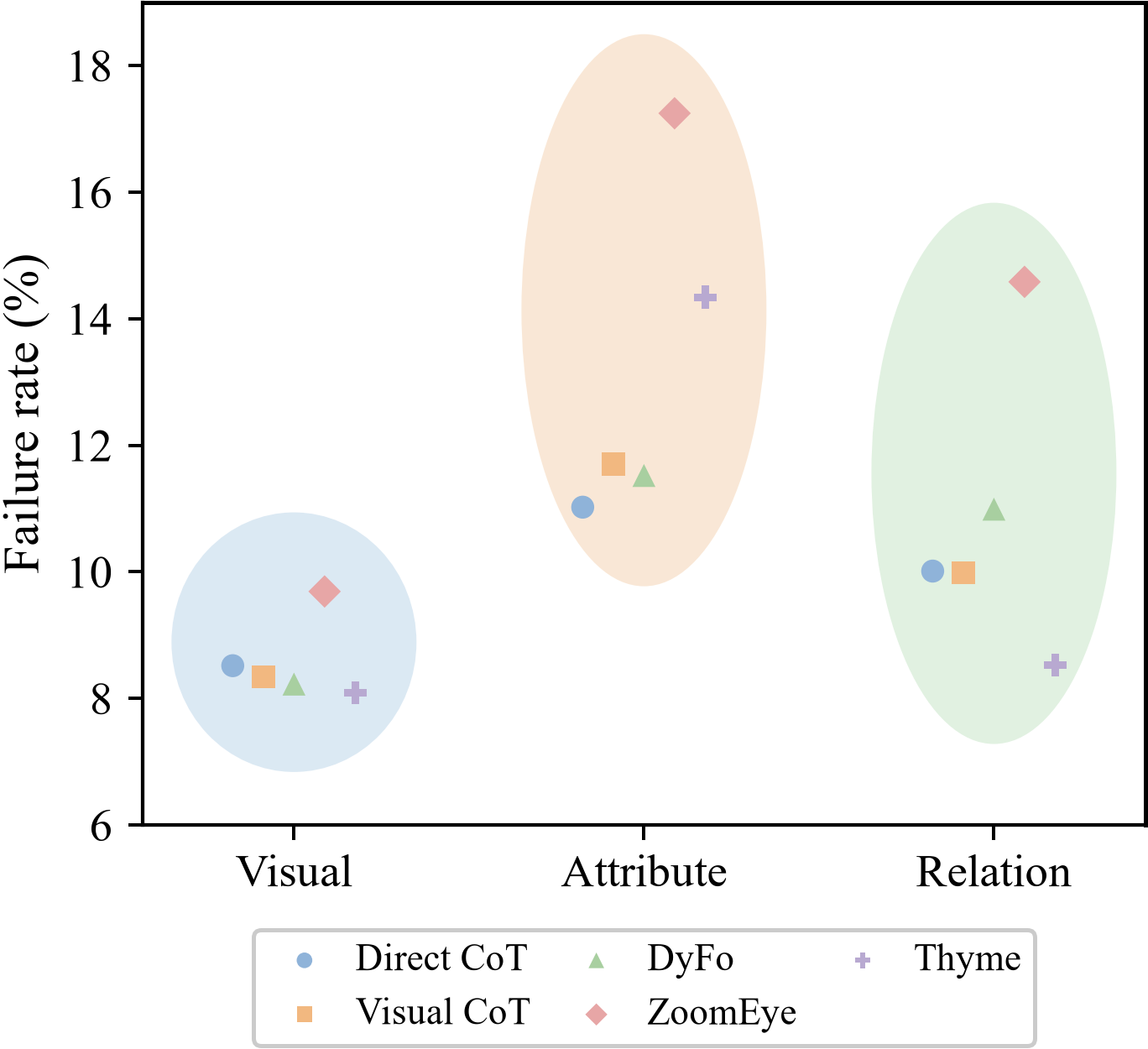}
        \caption{MAMI}
        \label{fig:category-7b-mami}
    \end{subfigure}
    \hfill
    \begin{subfigure}[t]{0.45\textwidth}
        \centering
        \includegraphics[width=\linewidth]{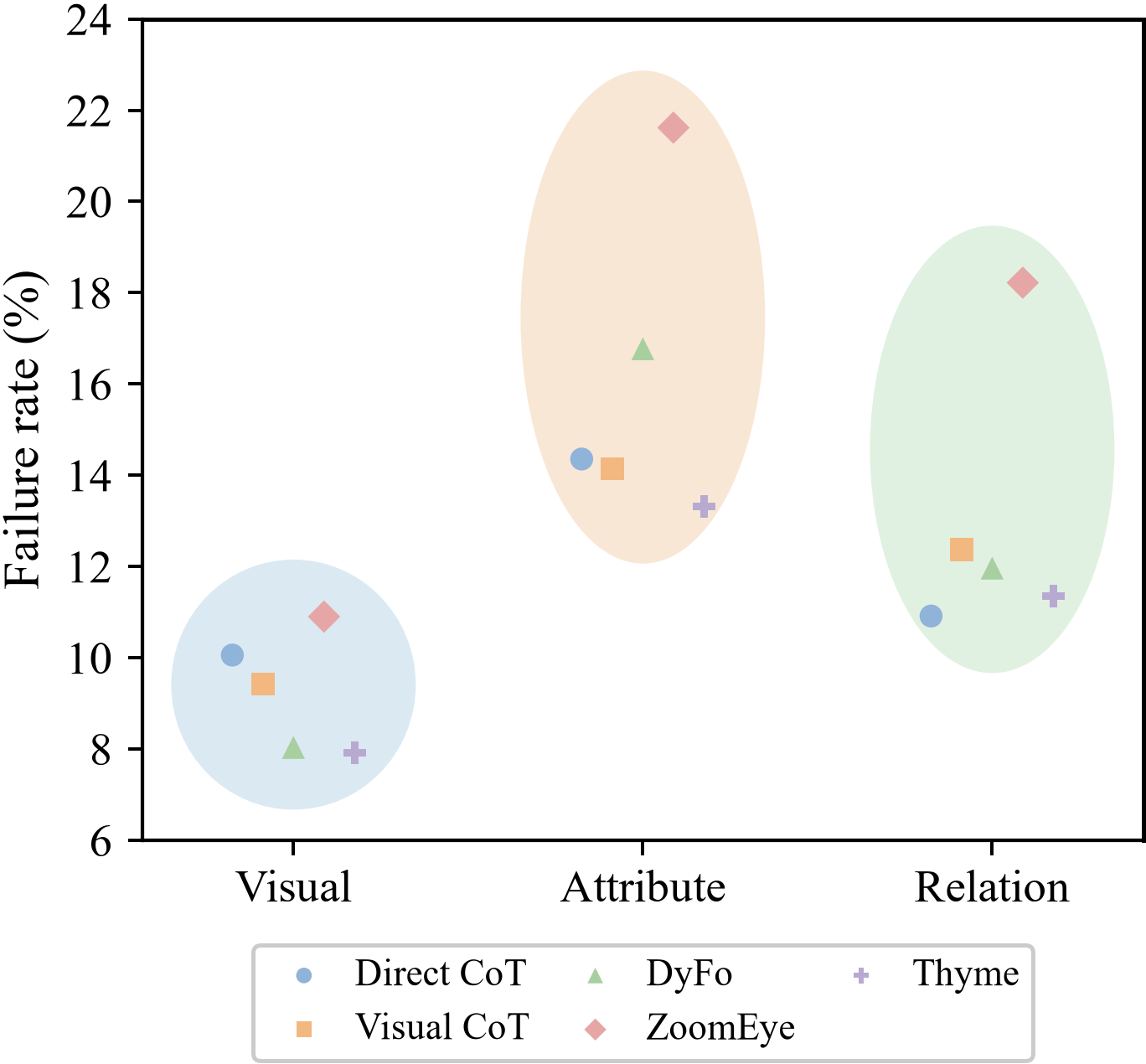}
        \caption{PrideMM}
        \label{fig:category-7b-pridemm}
    \end{subfigure}

    \caption{Atomic-fact failure rates on four datasets with Qwen2.5-VL-7B. Each marker denotes a rationale-generation method, and the shaded ellipses highlight the distribution within each category. Lower is better.}
    \label{fig:category-7b}
\end{figure*}

\begin{figure*}[t]
    \centering
    \begin{subfigure}[t]{0.45\textwidth}
        \centering
        \includegraphics[width=\linewidth]{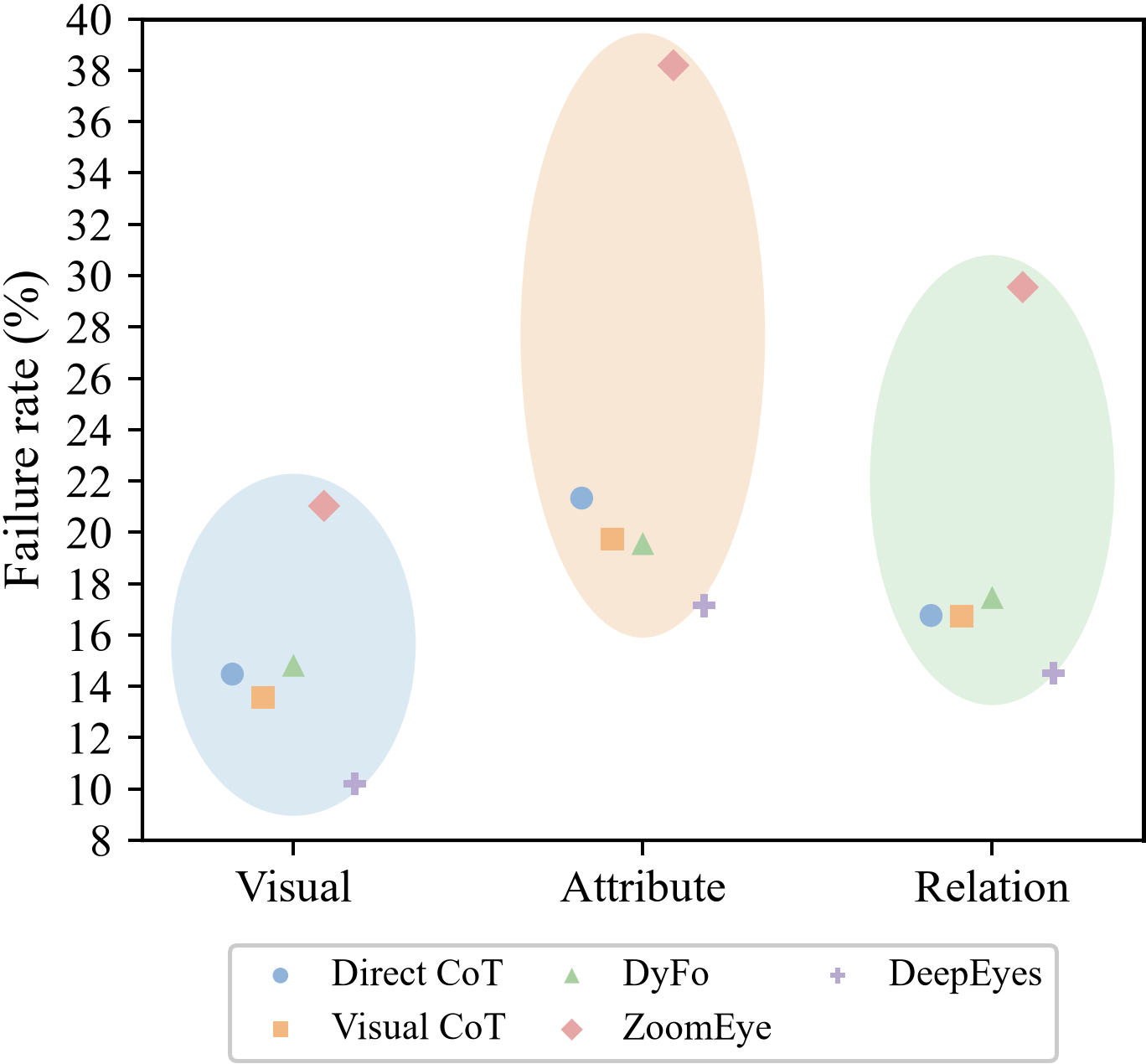}
        \caption{Harm-C}
        \label{fig:category-32b-harmc}
    \end{subfigure}
    \hfill
    \begin{subfigure}[t]{0.45\textwidth}
        \centering
        \includegraphics[width=\linewidth]{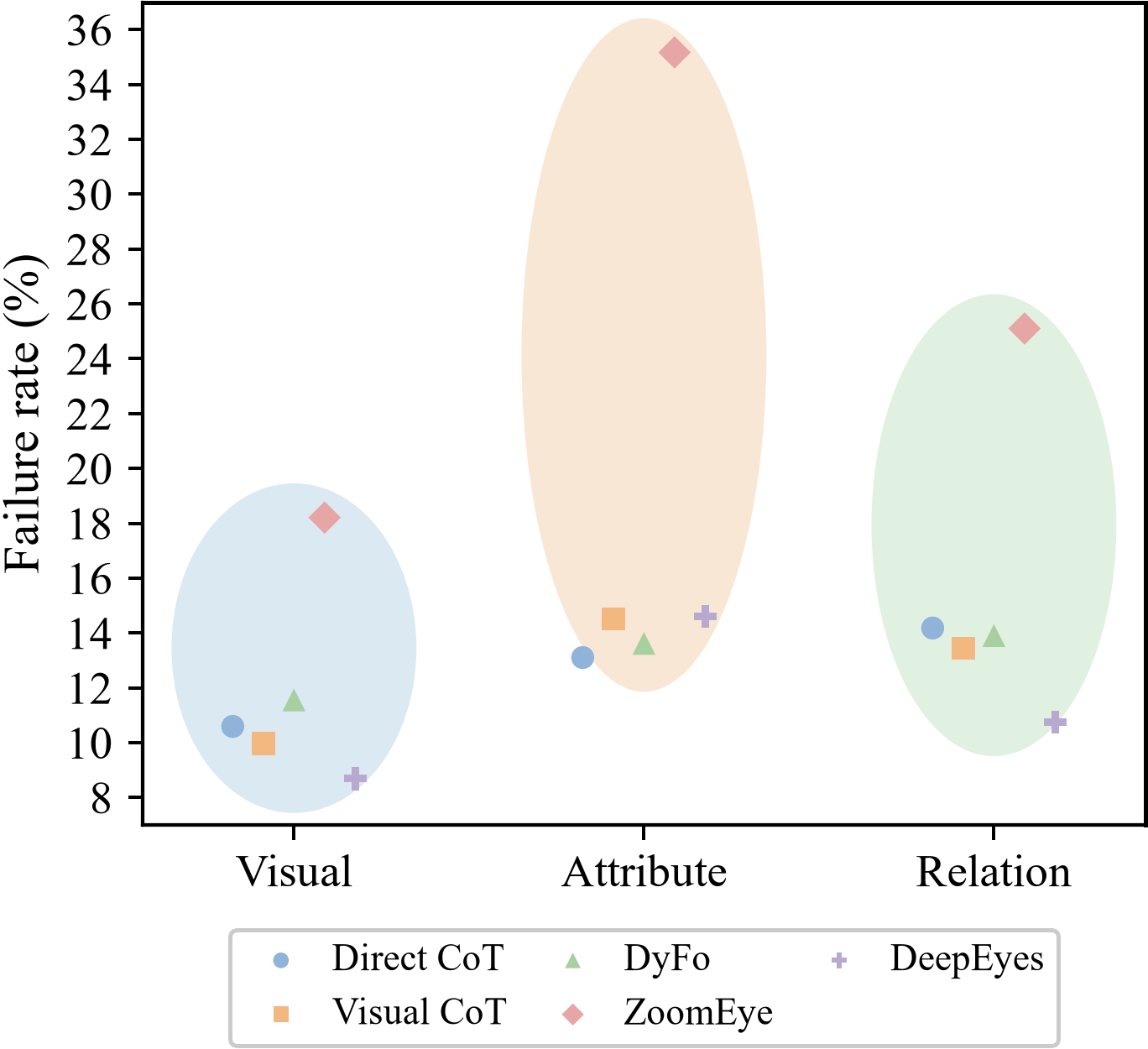}
        \caption{Harm-P}
        \label{fig:category-32b-harmp}
    \end{subfigure}

    \vspace{3pt}

    \begin{subfigure}[t]{0.45\textwidth}
        \centering
        \includegraphics[width=\linewidth]{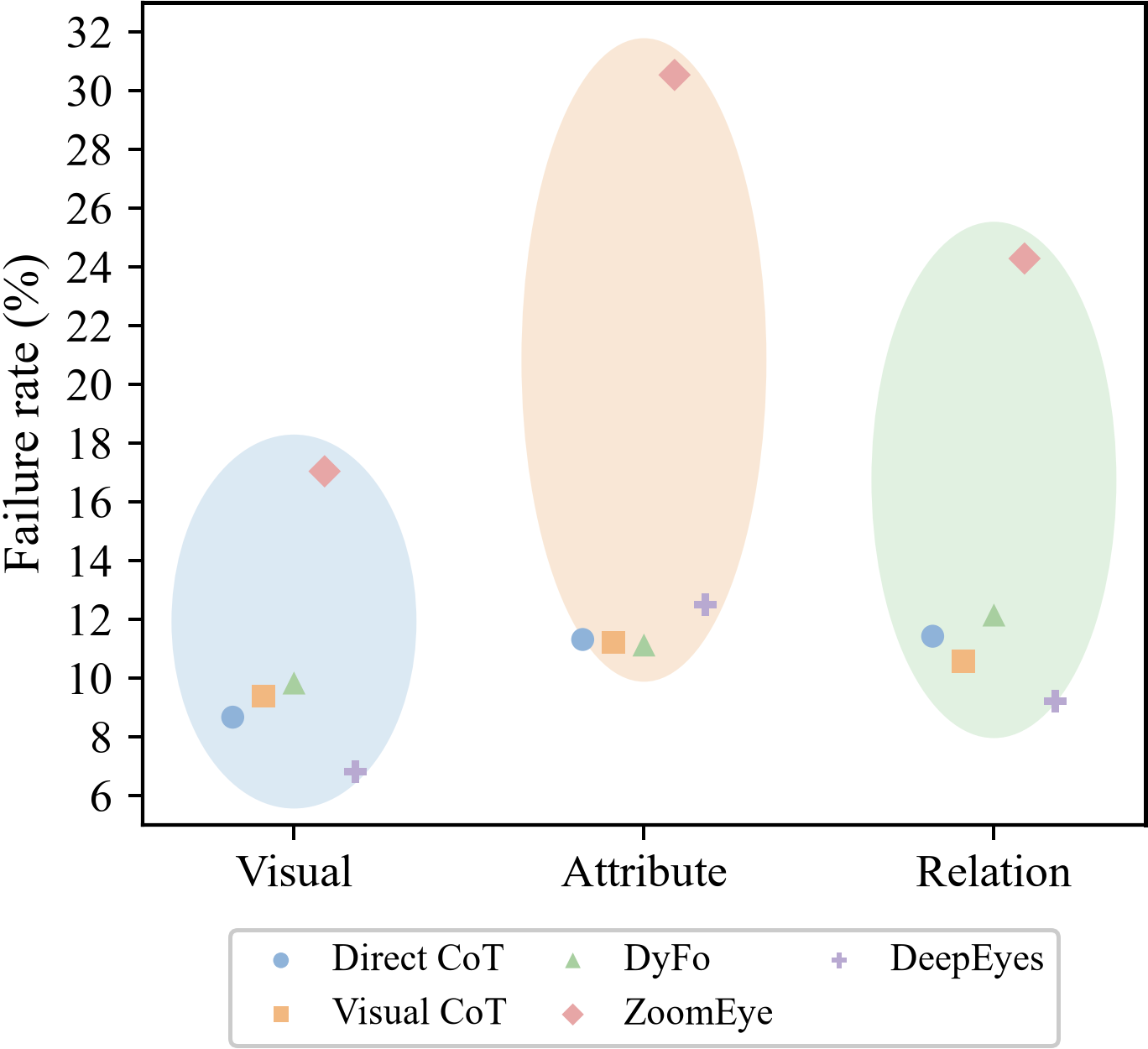}
        \caption{MAMI}
        \label{fig:category-32b-mami}
    \end{subfigure}
    \hfill
    \begin{subfigure}[t]{0.45\textwidth}
        \centering
        \includegraphics[width=\linewidth]{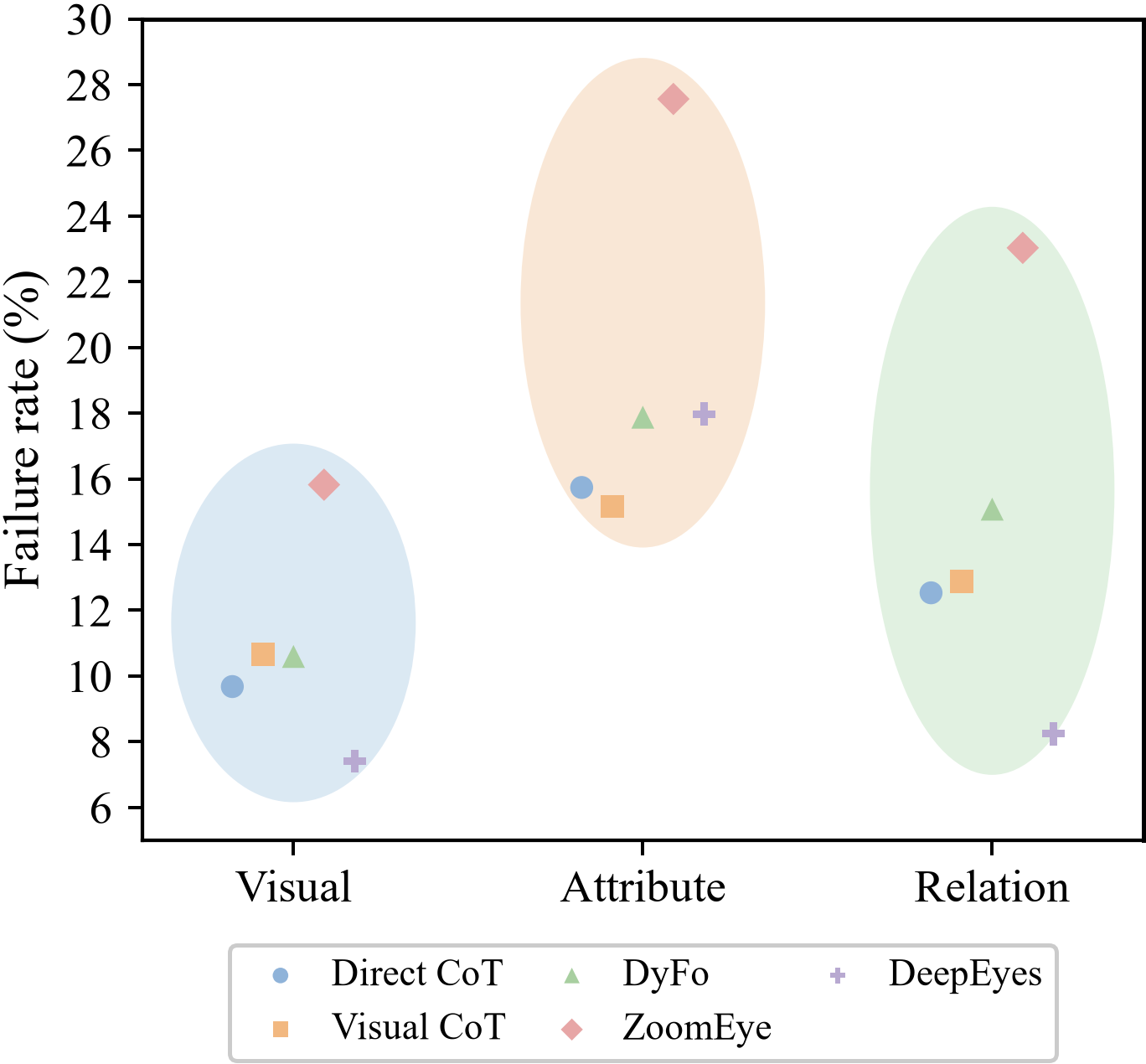}
        \caption{PrideMM}
        \label{fig:category-32b-pridemm}
    \end{subfigure}

    \caption{Atomic-fact failure rates on four datasets with Qwen2.5-VL-7B. Each marker denotes a rationale-generation method, and the shaded ellipses highlight the distribution within each category. Lower is better.}
    \label{fig:category-32b}
\end{figure*}

We provide additional results for the analysis in Section~\ref{sec:category_analysis}. 
Figures~\ref{fig:category-7b} and~\ref{fig:category-32b} report the failure rates of Qwen2.5-7B-VL and 
Qwen2.5-32B-VL, respectively, across Harm-C, Harm-P, MAMI, and PrideMM. 
The additional results are consistent with the observations in the main text: 
across datasets and methods, Attribute and Relation facts generally exhibit 
higher failure rates than Visual facts, indicating that models have greater 
difficulty faithfully verifying object properties and interactions than 
recognizing visible entities. This pattern holds for both model scales. Moreover, 
tool-augmented reasoning does not consistently reduce category-level 
hallucination. Visual CoT, DyFo, ZoomEye, and DeepEyes show mixed effects 
across datasets, and some tool-based methods even produce higher failure 
rates for Attribute or Relation facts. These results further support our 
conclusion that the availability or invocation of visual tools alone does 
not reliably improve image-grounded rationale generation.

\section{Case Study}
\label{app:case_study}

\subsection{hallucination case study}
\begin{figure*}[t]
    \centering
    \includegraphics[width=0.95\textwidth]{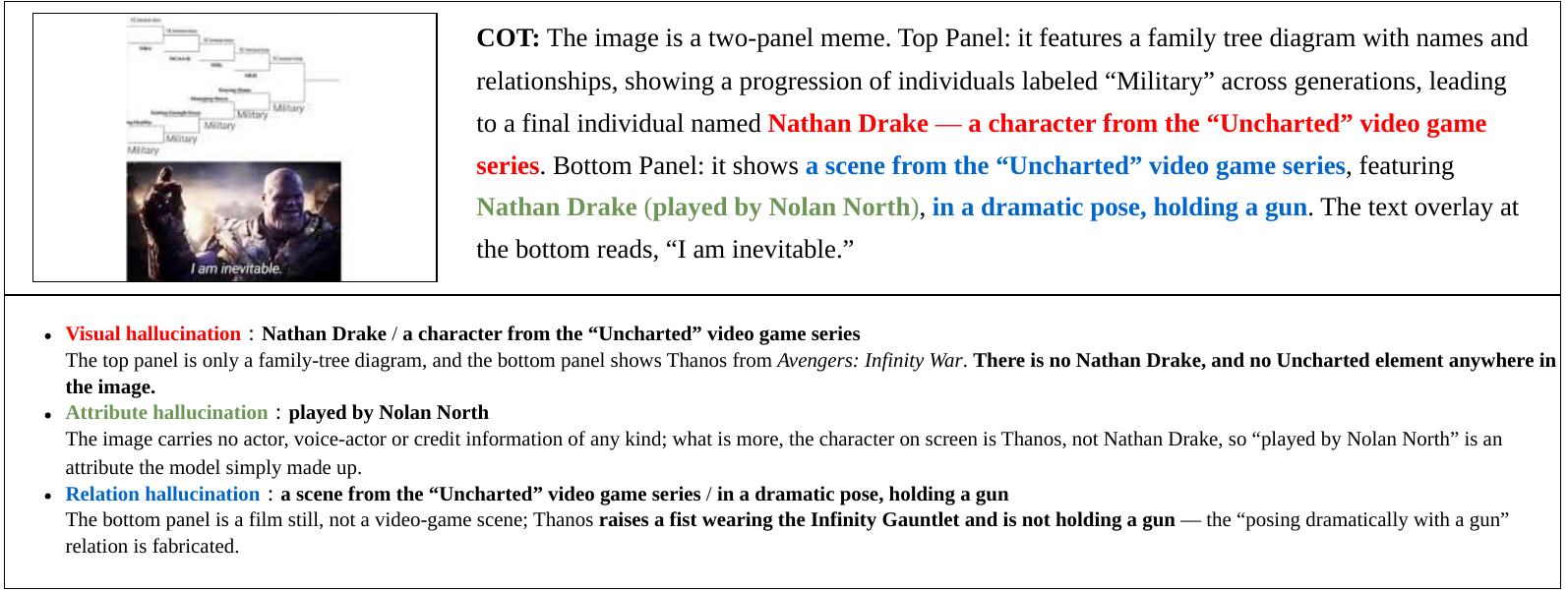}\\[2pt]
    \includegraphics[width=0.95\textwidth]{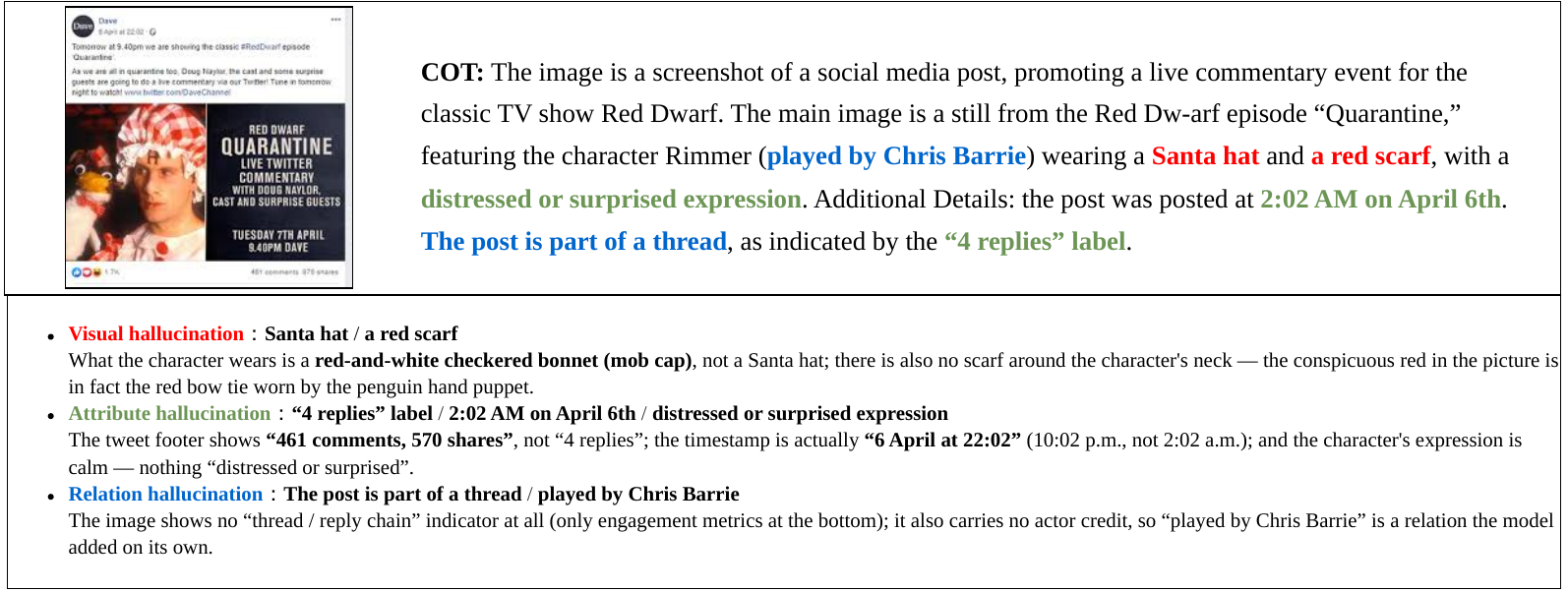}\\[2pt]
    \includegraphics[width=0.95\textwidth]{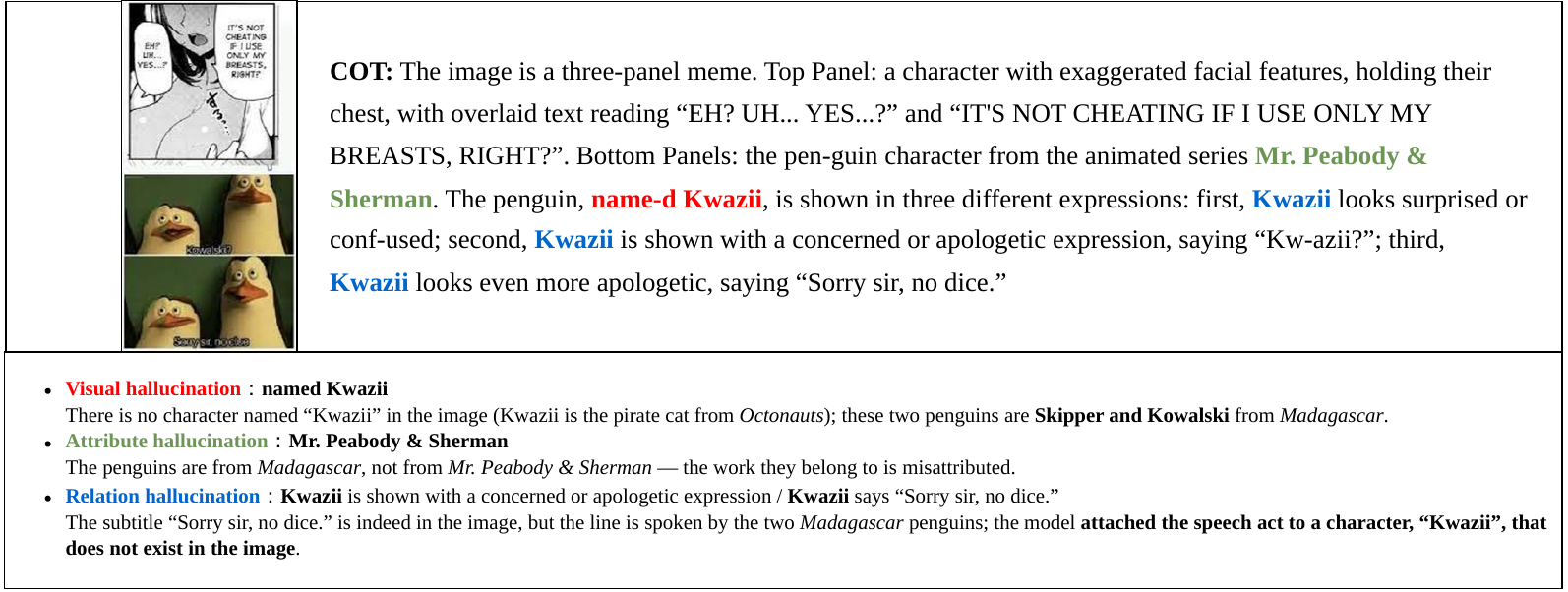}\\[2pt]
    \includegraphics[width=0.95\textwidth]{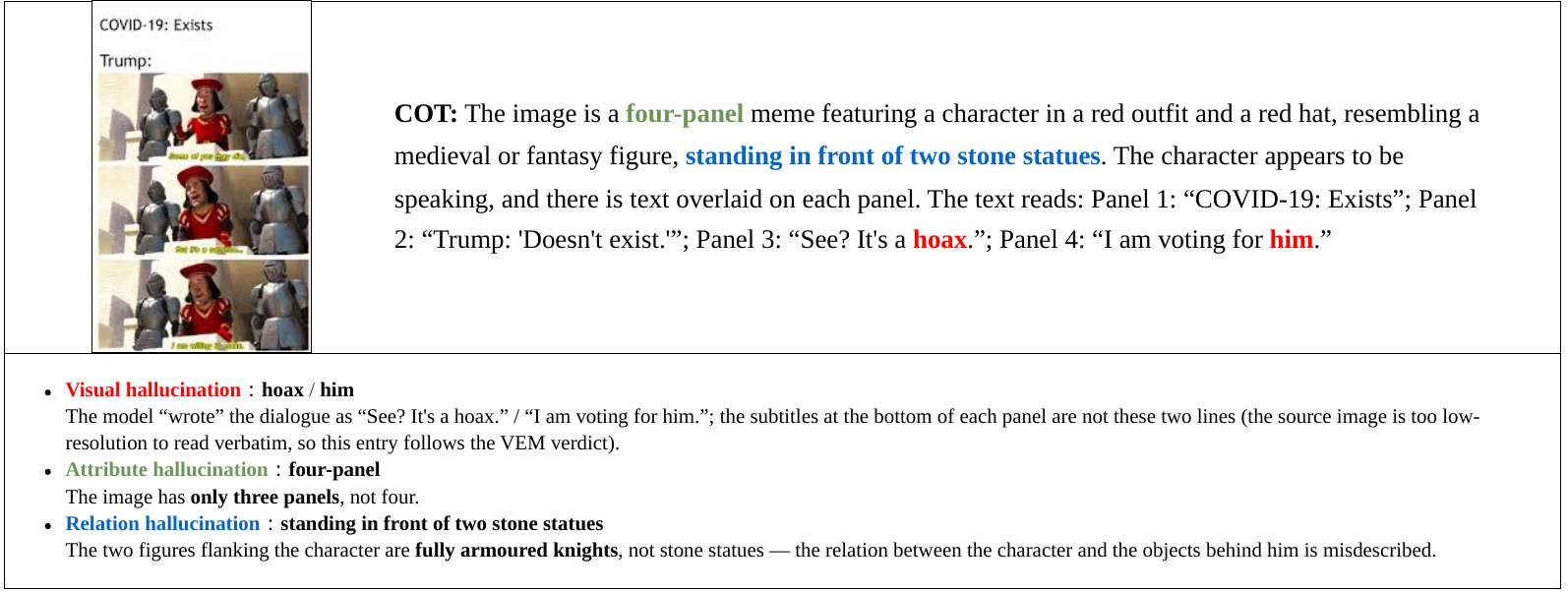}
    \caption{Representative examples of visual evidence hallucination. Each
    panel contains unsupported visual claims in the generated rationale,
    including hallucinated entities, attributes, or relations.}
    \label{fig:visual_case_study}
\end{figure*}

\begin{figure*}[t]
    \centering
    \begin{subfigure}[t]{0.48\textwidth}
        \centering
        \includegraphics[width=\linewidth]{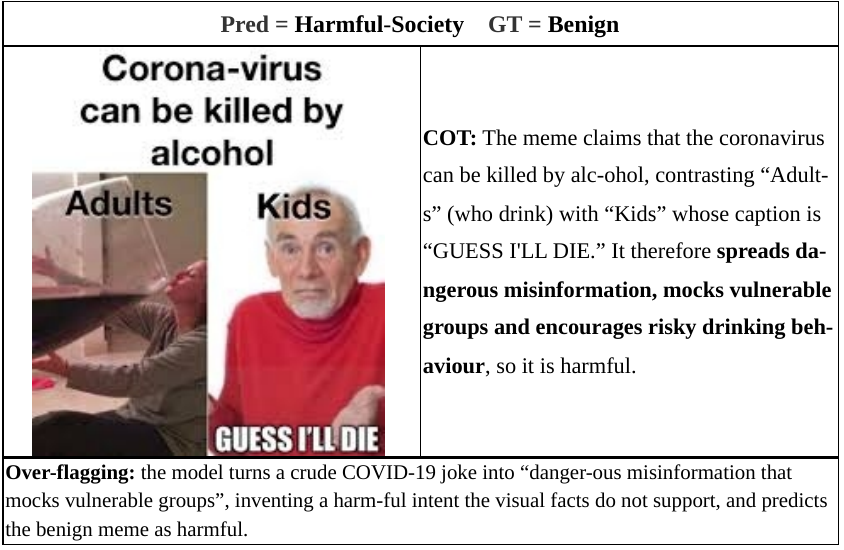}
        \label{fig:of_case_1}
    \end{subfigure}
    \hfill
    \begin{subfigure}[t]{0.48\textwidth}
        \centering
        \includegraphics[width=\linewidth]{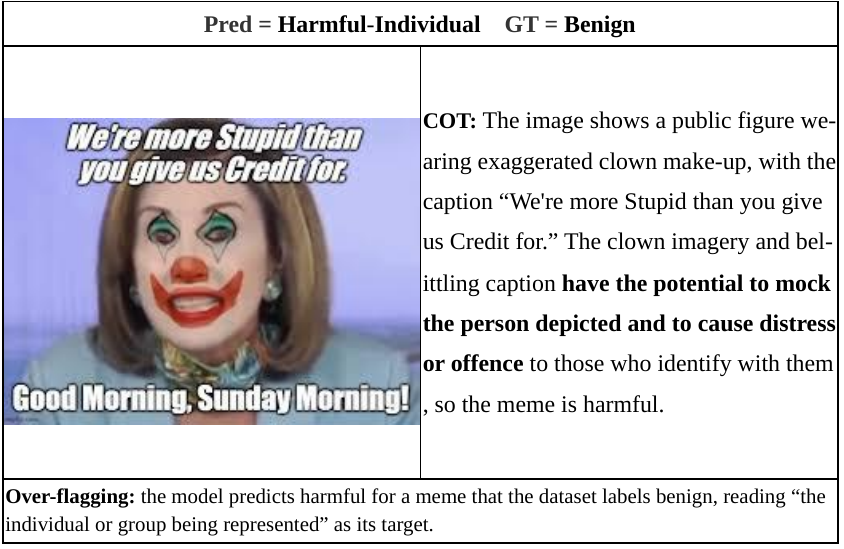}
        \label{fig:of_case_2}
    \end{subfigure}

    \vspace{3pt}

    \begin{subfigure}[t]{0.48\textwidth}
        \centering
        \includegraphics[width=\linewidth]{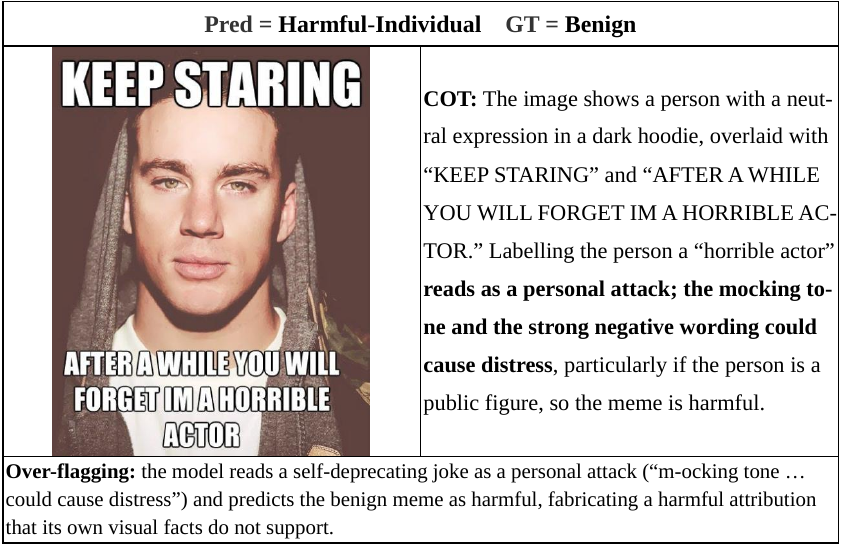}
        \label{fig:of_case_3}
    \end{subfigure}
    \hfill
    \begin{subfigure}[t]{0.48\textwidth}
        \centering
        \includegraphics[width=\linewidth]{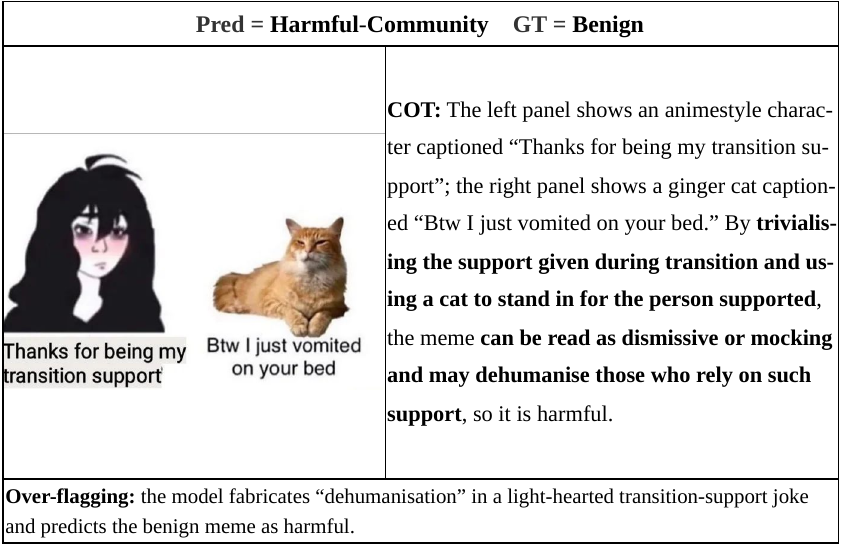}
        \label{fig:of_case_4}
    \end{subfigure}

    \caption{Representative examples of over-flagging (OF). The model assigns
    unsupported harmful attributions to benign image--text pairs without visual hallucination.}
    \label{fig:of_case_study}
\end{figure*}

\begin{figure*}[t]
    \centering
    \begin{subfigure}[t]{0.48\textwidth}
        \centering
        \includegraphics[width=\linewidth]{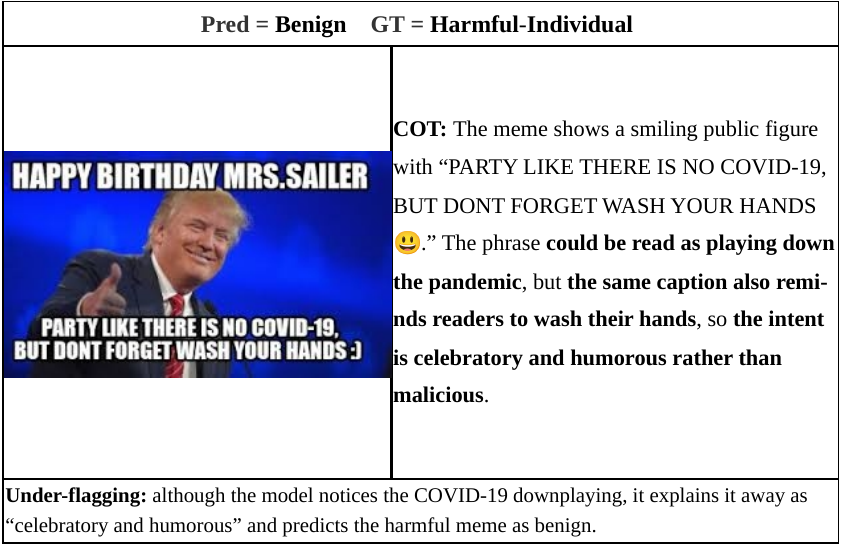}
        \label{fig:uf_case_1}
    \end{subfigure}
    \hfill
    \begin{subfigure}[t]{0.48\textwidth}
        \centering
        \includegraphics[width=\linewidth]{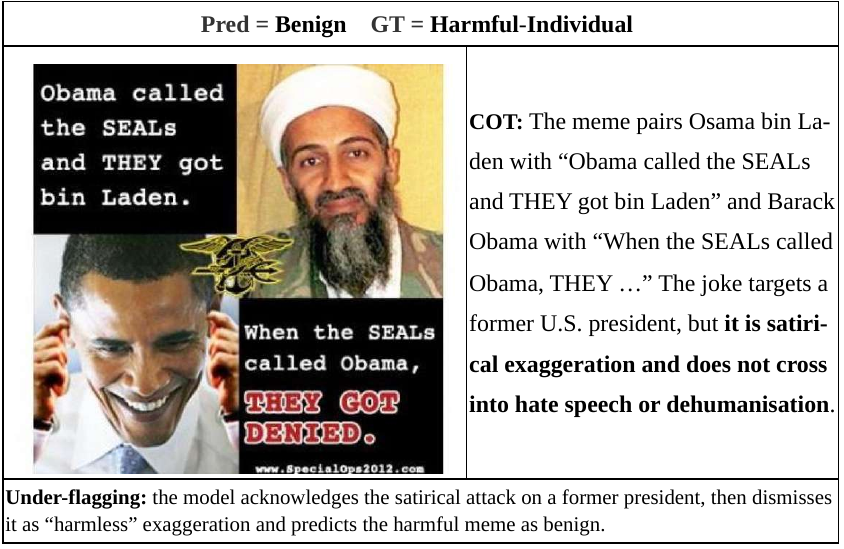}
        \label{fig:uf_case_2}
    \end{subfigure}

    \vspace{3pt}

    \begin{subfigure}[t]{0.48\textwidth}
        \centering
        \includegraphics[width=\linewidth]{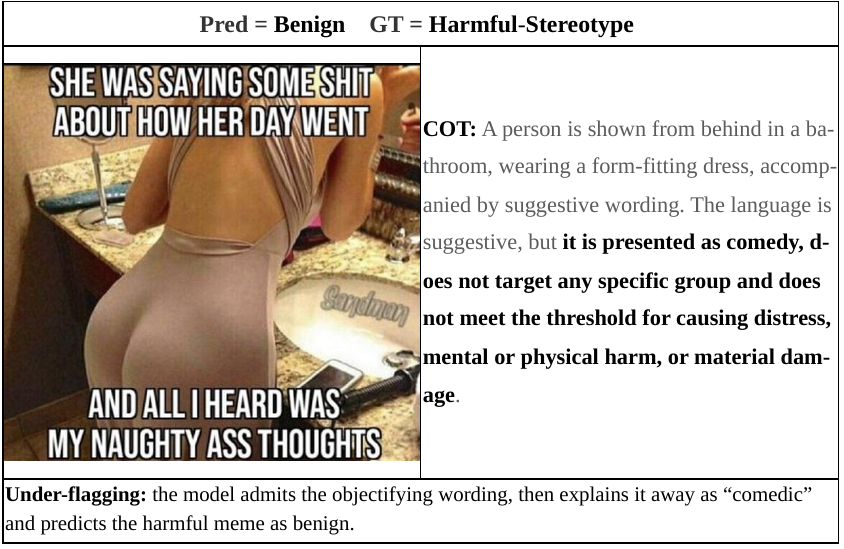}
        \label{fig:uf_case_3}
    \end{subfigure}
    \hfill
    \begin{subfigure}[t]{0.48\textwidth}
        \centering
        \includegraphics[width=\linewidth]{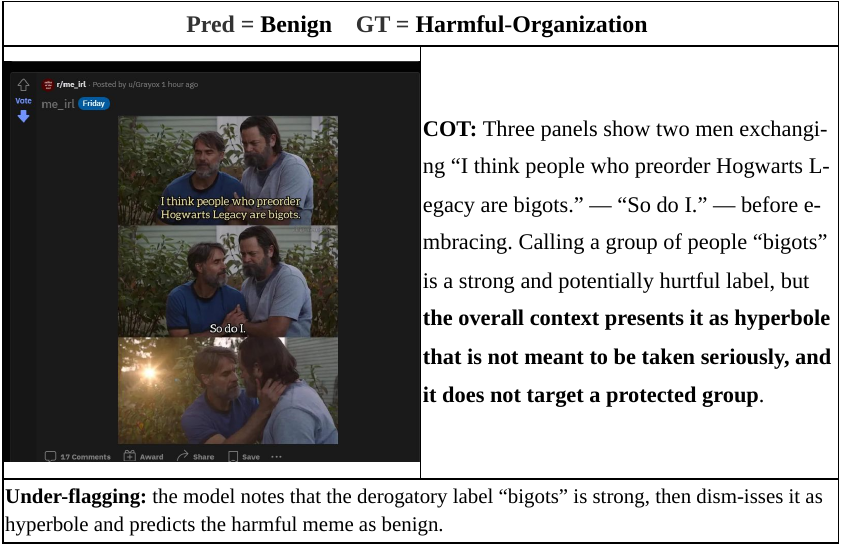}
        \label{fig:uf_case_4}
    \end{subfigure}

    \caption{Representative examples of under-flagging (UF). The model fails
    to identify a harmful attribution in harmful content and instead explains
    the image--text pair as benign, harmless, or merely humorous without visual hallucination.}
    \label{fig:uf_case_study}
\end{figure*}

\begin{figure*}[t]
    \centering
    \begin{subfigure}[t]{0.48\textwidth}
        \centering
        \includegraphics[width=\linewidth]{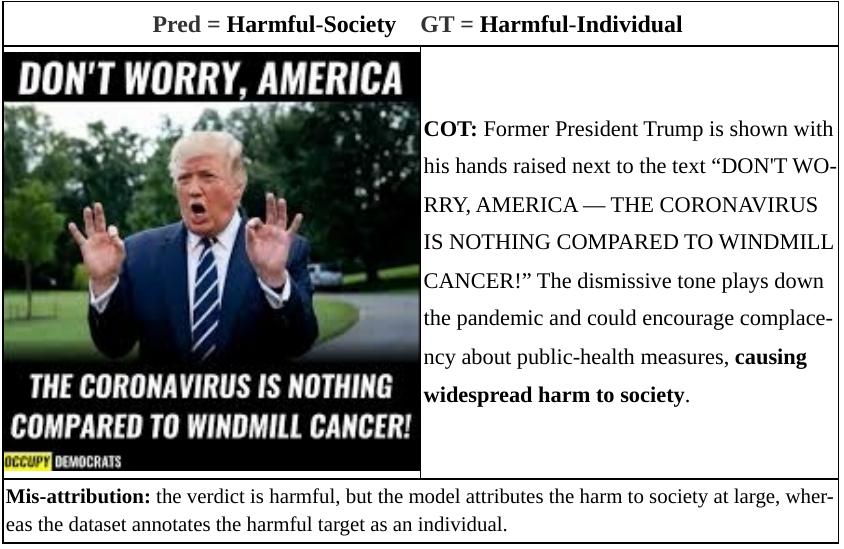}
        \label{fig:ma_case_1}
    \end{subfigure}
    \hfill
    \begin{subfigure}[t]{0.48\textwidth}
        \centering
        \includegraphics[width=\linewidth]{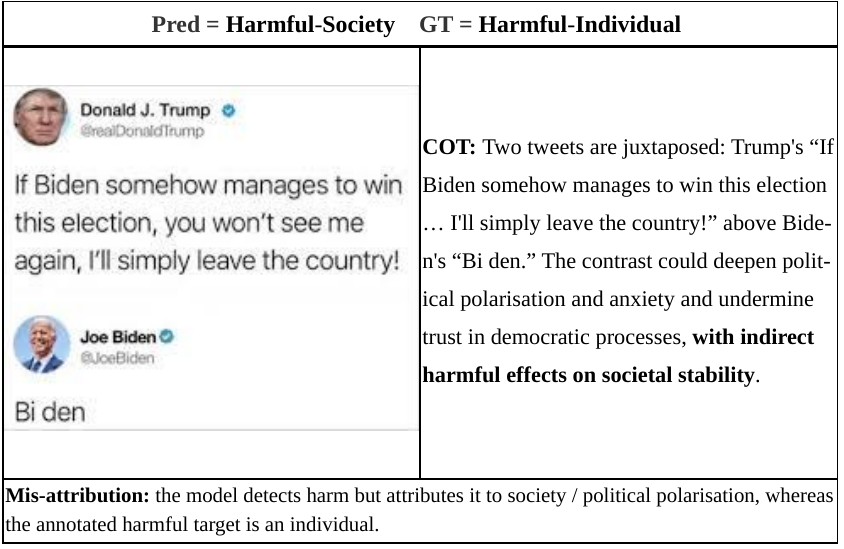}
        \label{fig:ma_case_2}
    \end{subfigure}

    \vspace{3pt}

    \begin{subfigure}[t]{0.48\textwidth}
        \centering
        \includegraphics[width=\linewidth]{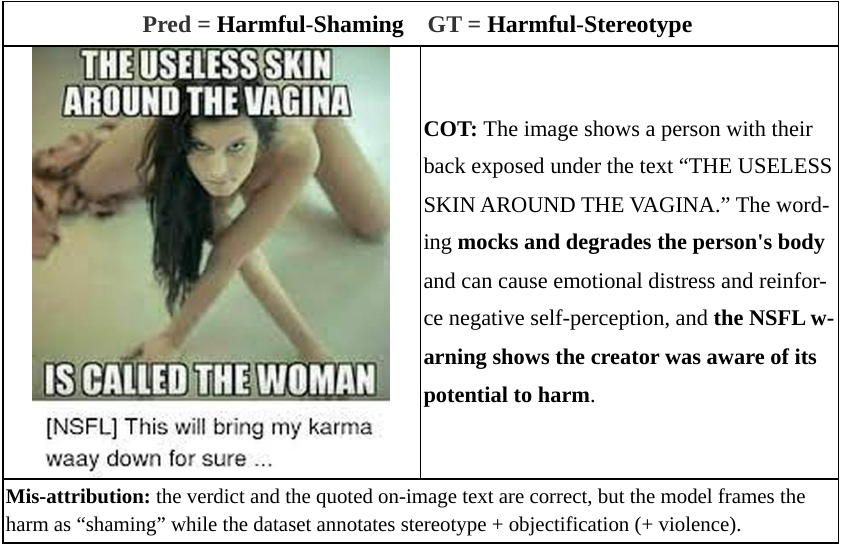}
        \label{fig:ma_case_3}
    \end{subfigure}
    \hfill
    \begin{subfigure}[t]{0.48\textwidth}
        \centering
        \includegraphics[width=\linewidth]{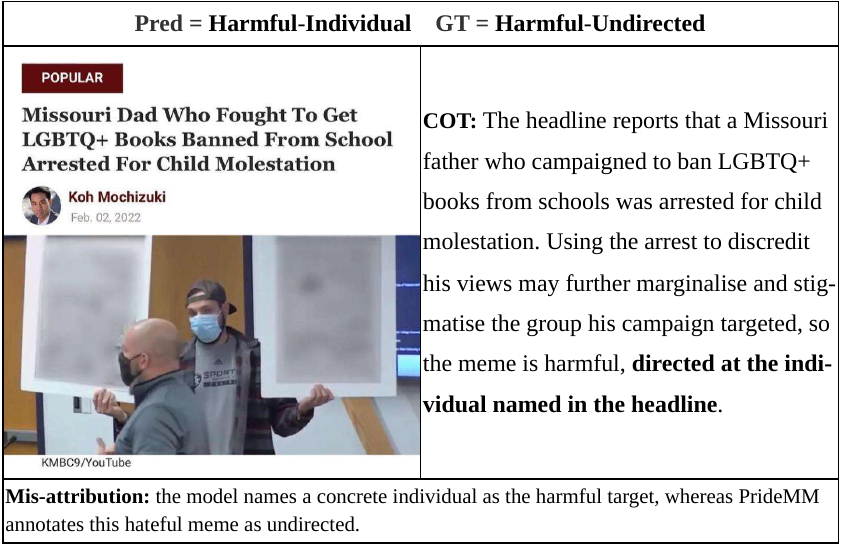}
        \label{fig:ma_case_4}
    \end{subfigure}

    \caption{Representative examples of mis-attribution (MA). The model
    correctly identifies the content as harmful but attributes the harmfulness
    to an incorrect target, category, or stance without visual hallucination.}
    \label{fig:ma_case_study}
\end{figure*}

To complement our quantitative evaluation, we present representative case
studies illustrating the three types of rationale hallucination investigated in
this work. Figure~\ref{fig:visual_case_study} shows four examples of visual
evidence hallucination, where the rationale contains visual claims that are not
supported by the input image. Figures~\ref{fig:of_case_study}--\ref{fig:ma_case_study}
present examples of safety attribution hallucination. Specifically,
over-flagging (OF) refers to assigning a harmful attribution to benign content,
under-flagging (UF) refers to failing to identify a harmful attribution in
harmful content, and mis-attribution (MA) refers to predicting the harmful
label correctly while assigning the wrong harmful target, category, or stance, as described in Section \ref{sec:metric}.
The examples demonstrate that a rationale may be fluent and superficially
plausible while still containing unsupported visual evidence or an incorrect
safety interpretation. For readability, the rationales in these case studies
are shortened and lightly edited from the original model outputs, while
preserving the corresponding hallucination type.

\subsection{TCG-RL Case Study}
\label{app:case_tcg}
\paragraph{Why TCG-RL improves grounding: qualitative evidence.} Our quantitative results in \ref{sec:tcg_results} show that TCG-RL substantially reduces rationale hallucination, and the case studies below illustrate the mechanism behind this improvement. The analysis in \ref{sec:tool_analysis} showed that Think-with-Image safety guards invoke visual tools infrequently, and that even when a crop is taken it often retains a large fraction of the original image rather than isolating the safety-relevant region. Tables~\ref{tab:case_strict} and~\ref{tab:case_relaxed} contrast TCG-RL with its tool-free reference Thyme on two Harm-C memes, where we observe that TCG-RL's tool calls are more consistently directed at the region that carries the harm. In Table~\ref{tab:case_strict}, Thyme zooms into the caption bands and the product region, yet misidentifies the bleach bottle as "cough drops"; this unsupported visual fact leads it to predict \textsc{not\_harmful} for a meme that instructs the reader to drink bleach, whereas TCG-RL reads the same captions and predicts \textsc{harmful} correctly. In Table~\ref{tab:case_relaxed} the difference is spatial: Thyme devotes both of its zoom calls to the person's face and to the held object and never crops the caption, where the harmful text actually lies, while TCG-RL's single crop isolates exactly that caption and its rationale contains no unsupported visual fact (VES $1.000$ vs.\ $0.867$). These examples are consistent with the aggregate trend in Table~\ref{tab:tcg_results}, where TCG-RL attains the highest VES on all four benchmarks (e.g., $+23.1$ points over Thyme on Harm-C) while also improving SAS over Thyme on every benchmark, indicating that more focused tool use translates into fewer visual and safety attribution hallucinations without sacrificing prediction quality.

\newcommand{\hallu}[1]{\textcolor{red}{\textbf{#1}}}      
\newcommand{\attrib}[1]{\textcolor{blue}{\textbf{#1}}}    
\newcommand{\thickhline}{\noalign{\hrule height 1.2pt}}
\newcommand{\padimg}[1]{%
  \raisebox{0pt}[\dimexpr\height+6pt\relax][\dimexpr\depth+6pt\relax]%
    {\includegraphics[width=\linewidth]{#1}}}

\begin{table*}[t]
\centering
\small
\setlength{\tabcolsep}{16pt}
\renewcommand{\arraystretch}{1.6}
\begin{tabular}{>{\centering\arraybackslash}m{0.42\textwidth}>{\centering\arraybackslash}m{0.42\textwidth}}
\thickhline
\textbf{Thyme} & \textbf{TCG-RL} \\
\hline
\padimg{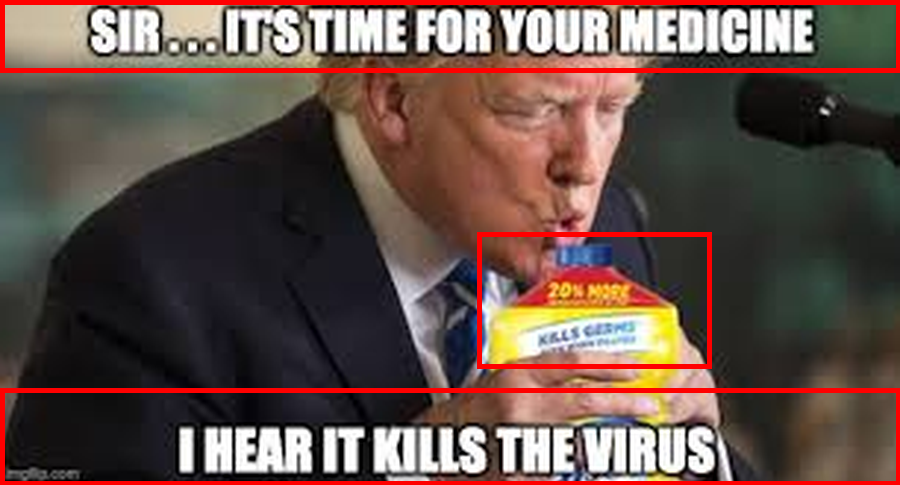}
&
\padimg{figs/table_assets/case1_000120_original_annotated.png}
\\[10pt]
\hline
\padimg{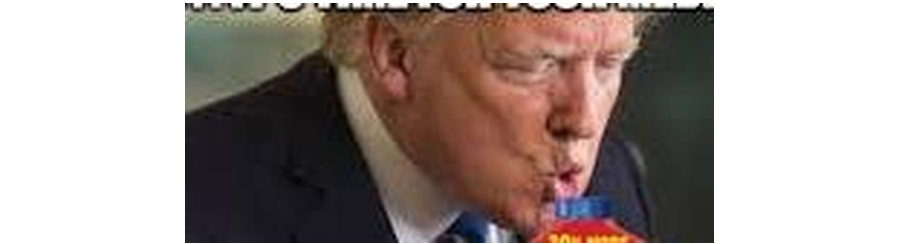}
&
\padimg{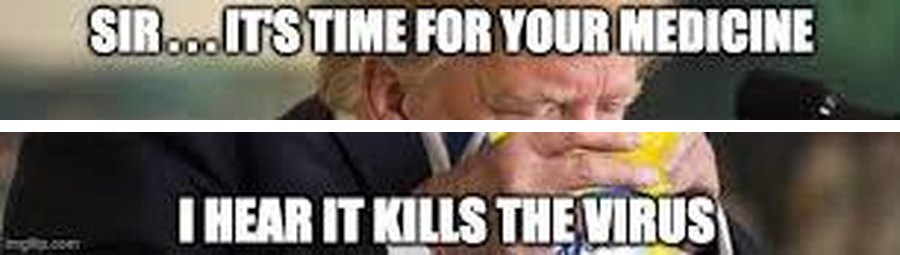}
\\[10pt]
\hline
{\footnotesize\raggedright
The meme features a man holding a container of \hallu{cough drops} with overlaid
text. The text at the top reads, ``SIR\ldots IT'S TIME FOR YOUR MEDICINE,'' and
the text at the bottom reads, ``I HEAR IT KILLS THE VIRUS.'' To ensure
there are no hidden harmful elements, I will crop the image to focus on the
overlaid text and the man's face. [TOOL\_USE] \hallu{The face crop shows a man
holding a container of cough drops.} There is no hate speech,
harassment, or misinformation present. It is a benign political joke.\par}
&
{\footnotesize\raggedright
The image contains overlaid text, which requires closer inspection to determine
the context and intent behind the meme. I need to zoom in on the overlaid text
to read it clearly and understand the message. [TOOL\_USE] After examining the
overlaid text more closely, the top caption reads ``SIR\ldots IT'S TIME FOR YOUR
MEDICINE,'' and the bottom caption says ``I HEAR IT KILLS THE VIRUS.'' This meme
appears to be making a humorous reference to a political figure and a product,
implying a connection between the two in a way that could be interpreted as
mocking or trivializing a serious health issue. It has the potential to cause
distress or spread misinformation about the virus and its treatment.\par}
\\[10pt]
\hline
{\footnotesize\textbf{[VERDICT] NOT\_HARMFUL}\newline \textcolor{red}{(GT: HARMFUL)}}
&
{\footnotesize\textbf{[VERDICT] HARMFUL}\newline \textcolor{green!60!black}{(GT: HARMFUL)}}
\\[6pt]
\thickhline
\end{tabular}
\caption{\textbf{Case study (\texttt{harmc\_test\_000120}).} Red boxes mark the
safety-relevant evidence and red spans the atomic facts judged unfaithful by the
FaithScore verifier. Thyme \hallu{misidentifies the bleach bottle as ``cough
drops''} and predicts \textsc{not harmful}, whereas TCG-RL predicts
\textsc{harmful} correctly.}
\label{tab:case_strict}
\end{table*}

\begin{table*}[t]
\centering
\small
\setlength{\tabcolsep}{16pt}
\renewcommand{\arraystretch}{1.6}
\begin{tabular}{>{\centering\arraybackslash}m{0.42\textwidth}>{\centering\arraybackslash}m{0.42\textwidth}}
\thickhline
\textbf{Thyme} & \textbf{TCG-RL} \\
\hline
\padimg{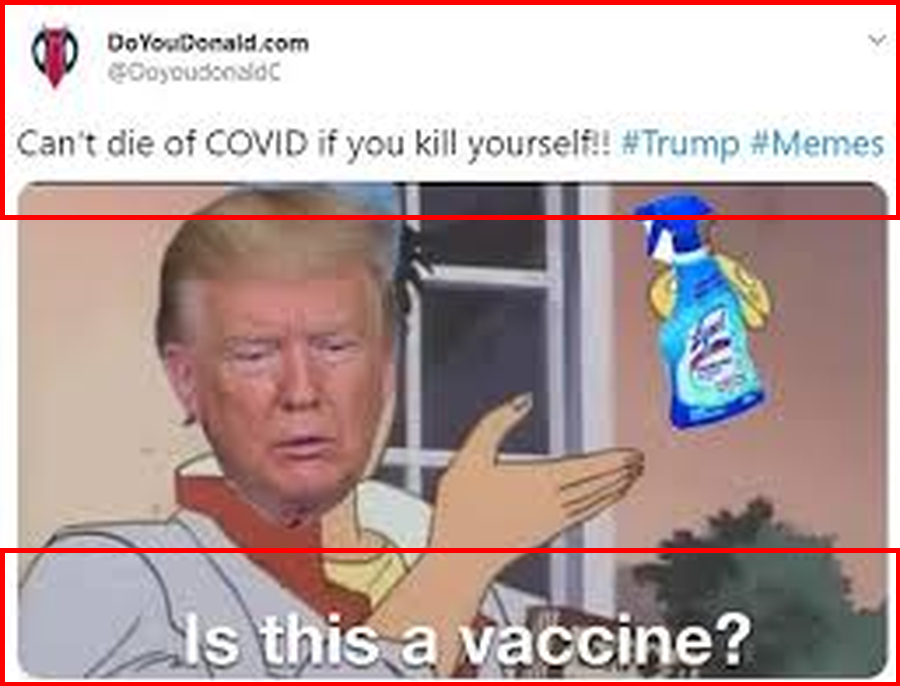}
&
\padimg{figs/table_assets/case2_000151_original_annotated.png}
\\[10pt]
\hline
\padimg{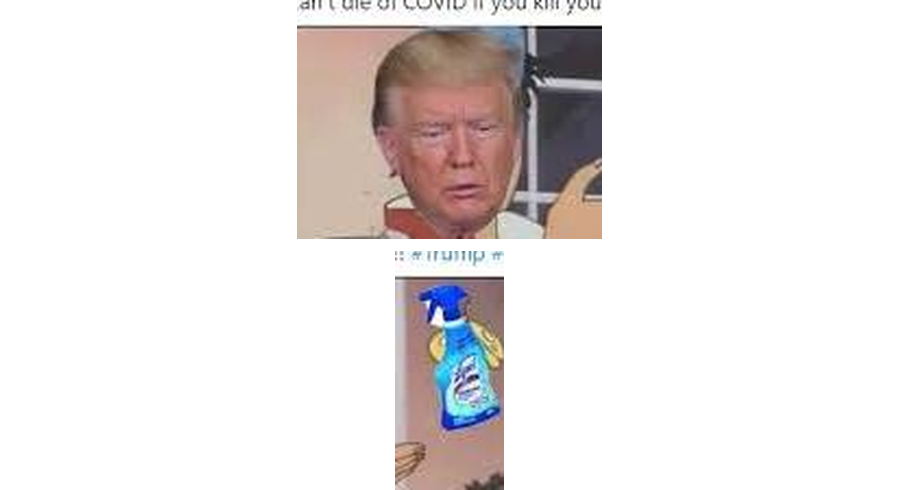}
&
\padimg{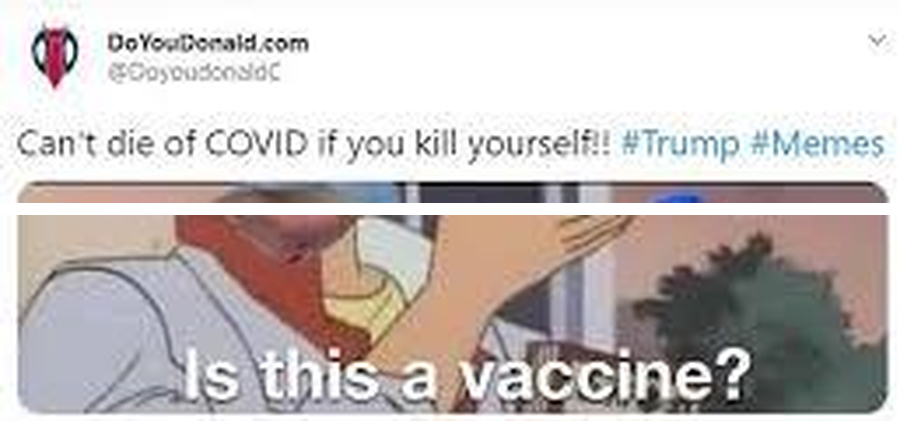}
\\[10pt]
\hline
{\footnotesize\raggedright
The meme contains overlaid text and a face, so I need to inspect the details
more closely. The text is already readable, but the face and the object it's
interacting with need to be examined. [TOOL\_USE] The face is that of a political
figure, and the object is \hallu{a bottle of hand sanitizer}. The overlaid text
reads, ``Can't die of COVID if you kill yourself!! Trump Memes'' and ``Is this a
vaccine?'' The combination of the text and the image suggests a harmful message,
as it implies that the hand sanitizer is a vaccine. Therefore, the meme
is harmful.\par}
&
{\footnotesize\raggedright
The image contains overlaid text, which requires closer inspection to determine
the context and intent behind the meme. I need to zoom in on the overlaid text
to read it clearly and understand the message. [TOOL\_USE] After examining the
overlaid text more closely, the top caption reads ``Can't die of COVID if you
kill yourself!! Trump Memes.'' This caption appears to be mocking a serious
health issue, which could be considered insensitive and potentially harmful to
individuals affected by COVID-19. The bottom caption adds to the context, but
the top caption is the primary concern here. Therefore, the meme is
harmful due to its potential to cause distress and spread misinformation.\par}
\\[10pt]
\hline
{\footnotesize\textbf{[VERDICT] HARMFUL}\newline \textcolor{green!60!black}{(GT: HARMFUL)}}
&
{\footnotesize\textbf{[VERDICT] HARMFUL}\newline \textcolor{green!60!black}{(GT: HARMFUL)}}
\\[6pt]
\thickhline
\end{tabular}
\caption{\textbf{Case study (\texttt{harmc\_test\_000151}).} Red boxes mark the
safety-relevant evidence, which lies in the caption alone, and red spans the
atomic facts judged unfaithful. Thyme zooms on the face and the spray bottle,
\hallu{misreads the spray as ``hand sanitizer''}; TCG-RL zooms on the caption
and produces no unfaithful fact.}
\label{tab:case_relaxed}
\end{table*}

\end{document}